\documentclass[sigconf]{acmart}

\AtBeginDocument{%
  }

\setcopyright{acmlicensed}
\acmYear{2025}

\acmConference[WWW '25
Companion]{Companion Proceedings of the ACM Web Conference 2025 }{April 28 - May 02, 2025}{Sydney, Australia}

\usepackage{algorithm}
\usepackage{algorithmic}

\usepackage{fancyhdr}
\usepackage{color}
\usepackage{url}
\usepackage{multirow}
\usepackage{graphicx} %
\usepackage{times}
\usepackage{latexsym}
\usepackage{multicol}
\usepackage{multirow}

\usepackage{placeins}

\newcommand{\methods}{{HPDT}\space}
\newcommand{\method}{{HPDT}}
\newcommand{\methodap}{{HPDT} wo G}
\newcommand{\methodgp}{{HPDT} wo A}
\newcommand{\methodt}{{HPDT} wo T}

\newcommand{\tokeng}{{\emph{global token\ }}}
\newcommand{\tokena}{{\emph{adaptive tokens\ }}}

\def\mse{\texttt{MSE}}
\def\gelu{\texttt{GELU}}
\def\knn{\texttt{KNN}}
\def\sample{\texttt{Segment}^z}
\def\TimetoVec{\texttt{Time2Vec}}
\def\DataMJ{\textsc{MuJoCo\ }}
\def\DataMW{\textsc{MetaWorld\ }}

\def\CVel{\textsc{Cheetah-Vel}}
\def\CDir{\textsc{Cheetah-Dir}}
\def\ADir{\textsc{Ant-Dir}}
\def\Reach{\textsc{Reach}}
\def\Pick{\textsc{Pick\&Place}}
\def\Hopper{\textsc{Param-Hopper}}
\def\Walker{\textsc{Param-Walker}}
\def\OMRL{OMRL}

\usepackage{tabularx} %
\usepackage{amsmath}
\usepackage{mathtools}
\usepackage{pifont}%

\usepackage{ragged2e}
\usepackage{booktabs}
\usepackage{xcolor}

\usepackage{colortbl}

\newcommand{\band}{\rowcolor{gray!20}}

\usepackage{url}

\newcommand{\cref}[1]{Condition~(\ref{#1})}

\usepackage{pifont} %
\usepackage{graphicx,subfigure,epsfig,fancybox} %
\usepackage{float}
\usepackage{color} %
\usepackage{multirow}

\newcommand{\revise}[1]{\textcolor{blue}{}}
\newcommand{\revised}[1]{\textcolor{blue}{}}

\usepackage{adjustbox}
\usepackage{array}
\usepackage{booktabs}

\newcommand{\vsiont}[1]{}

\DeclareMathAlphabet{\mathsfit}{\encodingdefault}{\sfdefault}{m}{sl}
\SetMathAlphabet{\mathsfit}{bold}{\encodingdefault}{\sfdefault}{bx}{n}

\def\mdpi{\mathcal{M}_i}

\def\pipolicyi{\pi_{\mathcal{M}_i}}

\def\Dseti{\mathcal{D}_{\mathcal{M}_i}}
\def\Dseto{\mathcal{O}_{\mathcal{M}_i}}
\def\Dsetoj{\mathcal{O}_{\mathcal{M}_j}}

\def\Prt{\text{Prom}}

\def\Trans{\mathcal{T}}
\def\init{\beta}

\def\Kerneli{\Trans_{\mathcal{M}_i}}
\def\Rewardi{\mathcal{R}_{\mathcal{M}_i}}

\def\Kernelj{\Trans_{\mathcal{M}_j}}
\def\Rewardj{\mathcal{R}_{\mathcal{M}_j}}

\def\conditiong{g_{\mathcal{M}_i}^z}
\def\conditionge{g_{\mathcal{M}_j}^z}

\def\mg{\mathbf{h}_{\theta_g}}
\def\ma{\mathbf{h}_{\theta_a}}

\def\msigma{\mathbf{\sigma}}
\def\myarrow{\leftrightarrows}

\def\mdpj{\mathcal{M}_j}

\def\pitargetj{\pi_{\mathcal{M}_j}^*}
\def\pipolicyj{\pi_{\mathcal{M}_j}}

\def\Dsetj{\mathcal{D}_{\mathcal{M}_j}}

\def\Prob{\mathcal{P}}

\def\mdp{\mathcal{M}}
\def\pitarget{\pi_{\mathcal{M}}^*}

\def\Sset{S}
\def\Aset{A}

\def\Reward{\mathcal{R}}
\def\Rtg{\hat{r}}

\def\defeq{:=}

\def\gB{{\mathcal{B}}}

\def\gL{{\mathcal{L}}}

\newcommand{\R}{\mathbb{R}}

\def\etal{{\it et~al.}}

\usepackage{enumitem}
\setlist[itemize]{leftmargin=*}

\usepackage[nolist,nohyperlinks]{acronym}

\makeatletter
\newcommand{\removelatexerror}{\let\@latex@error\@gobble}
\makeatother

\usepackage{url}
\usepackage{enumitem}

\include{macros}

\usepackage{enumitem}
\setlist[itemize]{leftmargin=*}

\usepackage{color}
\definecolor{colora}{rgb}{.7, .1, .1}

\begin{document}

\title[HPDT: Improving Few-Shot Policy Generalization with Global and Adaptive Guidance]{Hierarchical Prompt Decision Transformer: Improving Few-Shot Policy Generalization with Global and Adaptive Guidance}

\author{Zhe Wang}
\email{zw6sg@virginia.edu}
\affiliation{%
  \institution{University of Virginia}
  \city{Charlottesville}
  \state{Virginia}
  \country{USA}
}

\author{Haozhu Wang}
\email{haozhuw@amazon.com}
\affiliation{%
  \institution{Amazon}
  \country{USA}
}

\author{Yanjun Qi}
\email{yanjunqi@amazon.com}
\affiliation{%
  \institution{Amazon}
  \country{USA}
}
\renewcommand{\shortauthors}{Wang et al.}

\begin{abstract}

Decision transformers recast reinforcement learning as a conditional sequence generation problem, offering a simple but effective alternative to traditional value or policy-based methods. A recent key development in this area is the integration of prompting in decision transformers to facilitate few-shot policy generalization. However, current methods mainly use static prompt segments to guide rollouts, limiting their ability to provide context-specific guidance. Addressing this, we introduce a hierarchical prompting approach enabled by retrieval augmentation. Our method learns two layers of soft tokens as guiding prompts: (1) global tokens encapsulating task-level information about trajectories, and (2) adaptive tokens that deliver focused, timestep-specific instructions. The adaptive tokens are dynamically retrieved from a curated set of demonstration segments, ensuring context-aware guidance. Experiments across seven benchmark tasks in the \DataMJ and \DataMW environments demonstrate  the proposed approach consistently outperforms all baseline methods, suggesting that hierarchical prompting for decision transformers is an effective strategy to enable few-shot policy generalization.

\end{abstract}


\maketitle

\section{Introduction}

Recent literature includes two groups of methods for orchestrating operations to solve complex tasks: Large Language Model (LLM) orchestration \citep{yao2022react, shen2024hugginggpt} and Decision Transformers (DTs) \citep{chen2021decision}. LLM orchestration focuses on coordinating multiple language models or tools to tackle diverse, multi-faceted tasks. In contrast, DTs excel in environments with limited data and constrained computational resources by training a single transformer architecture on reward-conditioned control sequences. DTs specialize in offline reinforcement learning (RL)~\citep{levine2020offline,prudencio2023survey} by reframing policy learning as a sequence generation problem. Unlike the resource-intensive and broad capabilities of LLM orchestration, DTs provide a streamlined, efficient solution for sequential decision-making in structured, goal-conditioned scenarios.

DTs exhibit key advantages in capturing long-term dependencies, learning from offline datasets, and modeling return-conditioned policies, making them effective in domains such as robotics and autonomous systems. Empirical results by Chen~\etal\citep{chen2021decision} and Janner~\etal\citep{janner2021sequence} highlight their performance in sequence modeling and sample-efficient learning across diverse tasks. Moreover, theoretical work by Lin~\etal\citep{lin2023transformers} validates DTs’ ability to approximate the conditional expectation of expert policies. While LLM orchestration is versatile and powerful, DTs offer a focused and computationally efficient alternative for specific reinforcement learning challenges, complementing the broader applications of LLM-based systems.

Despite their strengths, DTs are inherently limited to individual task learning and struggle to adapt to new tasks without retraining. The Prompt Decision Transformer (PDT) \citep{xu2022prompting} addresses this limitation by incorporating demonstration-based prompts to guide action generation, enabling few-shot generalization. However, PDT faces challenges in optimally representing both global task identity and local context for adaptive decision-making. Its reliance on randomly sampled, static demonstration segments often leads to suboptimal guidance, particularly in Offline Meta-Reinforcement Learning (OMRL) settings~\citep{mitchell2021offline,li2021focal}, which demand effective adaptation strategies.

To overcome these limitations, we propose the Hierarchical Prompt Decision Transformer (\method), a novel approach inspired by hierarchical reinforcement learning (HRL) \citep{kahneman2011thinking, munkhdalai2017meta} and retrieval-augmented generation (RAG) \citep{lewis2020retrieval, goyal2022retrieval}. \method\ introduces a hierarchical prompt learning framework that enhances few-shot policy generalization by leveraging the structural information embedded in demonstration data. Specifically, \method\ learns two levels of soft prompt tokens:

\begin{enumerate}
    \item \tokeng, which encodes global task-level information (e.g., transition dynamics and rewards), and
    \item \tokena, which provides timestep-specific action guidance by retrieving relevant experiences from a demonstration set.
\end{enumerate}
This hierarchical prompt design enables dynamic and context-aware decision-making, seamlessly integrating global task understanding with local adaptability.

We summarize our key contributions as follows:
\begin{enumerate}
\item Hierarchical Prompt Framework: We introduce a novel hierarchical prompt learning framework to enhance Decision Transformers for improved few-shot policy generalization.
\item Dynamic and Context-Aware Guidance: \method\ employs optimized soft tokens to provide task-level (global) and timestep-specific (adaptive) prompting, facilitating stronger in-context guidance.
\item Empirical Validation: Extensive experiments demonstrate that \method\ achieves competitive performance across multiple environments, evaluated on benchmarks from both \DataMJ and \DataMW control domains.
\end{enumerate}

\section{Preliminaries}

\paragraph{Offline RL \& \OMRL.}\  RL task can be formalized as a Markov decision process $\mdp\defeq \langle\Sset, \Aset, \Reward, \Trans, \init \rangle$, which consists of a state space $\Sset$, an action space $\Aset$, a reward function $\Reward:\Sset\times\Aset\to\R$,  a transition dynamic $\Trans:\Sset\times\Aset\to\Sset$, and an initial state distribution $s_0 \sim \init$. A policy $\pi:\Sset\to\Aset$ interacts with the environment. At each time step $t \ge 0$, an action $a_t\sim\pi(s_t)$ is output by the policy $\pi$ and gets applied to the environment. After the agent performs action $a_t$, the environment transitions into the next state $s_{t+1}\sim\Trans(s_t,a_t)$ and produces a scalar reward $r_t\sim\Reward(s_t,a_t)$ as a feedback measuring the quality of the action $a_t$. The goal of RL is to learn an optimal policy $\pitarget$ that maximizes the accumulated reward within a time horizon $T$.\footnote{We focus on the environments with finite time horizon, but the definition generalizes to $T=\infty$. We also skip the constant discount factor for a better reading experience.}
\begin{align}
\vspace{-3mm}
\pitarget = \arg\max_{\pi} \sum_{t=0}^Tr_t.
\end{align}
In offline RL, the agent has no access to the environment during training and instead learns from a set of logged interaction trajectories, where each trajectory includes $\{s_{0}, a_{0}, r_{0}, s_{1}, a_{1}, r_{1},\cdots, s_{T}, a_{T},r_{T}\}.$ To approximate the optimal policy $\pitarget$, the static dataset should span as wide a distribution over $\Sset \times \Aset $ as possible~\citep{schweighofer2021understanding,kumar2022should}.

\OMRL\ targets at learning an agent that can efficiently approximate $\pitargetj$ for an unseen task $\mdpj$ when given few demonstrations of $\mdpj$ by learning-to-learn from multiple training tasks $\{\mdpi\}_{i=1}^n$ \footnote{Here we slightly abuse the notations, using $\mdp$ denotes both an RL task and its MDP, for few math notations.}. To achieve positive cross-task knowledge transfer, \OMRL\ assumes different tasks share the same state, same action space, and differ in their transition dynamics and the reward functions~\citep{yu2020meta}. 

We denote training tasks as a set:  $\{\mdpi\defeq \langle\Sset, \Aset, \Rewardi, \Kerneli, \init_{\mdpi} \rangle\}_{i=1}^n$. Similarly, we denote testing tasks as $\{\mdpj \defeq \langle\Sset, \Aset, \Rewardj, \Kernelj, \init_{\mdpj} \rangle\}_{j=1}^{n'}$. Every training task $\mdpi$ (or test task $\mdpj$) is associated with \textbf{a set of trajectory demonstrations } denoted as $\Dseti$ (or $\Dsetj$ for test task $\mdpj$) that includes only few, for instance, $5$ or $10$ historical interaction trajectories from this specific RL task. 
Each training task in $\{\mdpi\}_{i=1}^n$  associates with \textbf{a set of rollout trajectories} denoted as $\{\Dseto\}_{i=1}^n$, on which we train task-specific policies $\{\pipolicyi\}_{i=1}^n$. The derived task-specific policy $\pipolicyj$ is supposed to approximate $\pitargetj$ and can be evaluated by interacting with the environment. Different \OMRL\ algorithms
use training tasks' demonstration sets $\Dseti$ in different ways, though, all aiming to learning-to-learn the meta-knowledge across tasks and how efficiently tailoring the meta-knowledge to a specific task.

\paragraph{Decision Transformer (DT)}\  DT is a simple yet effective alternative to conventional value or policy-based methods for offline RL. It recasts the policy learning in offline RL as a sequence generation problem. Trajectory sequences from a static dataset are re-organized as $\{\Rtg_0, s_0, a_0, \Rtg_1, s_1, a_1\, \cdots, \Rtg_T, s_T, a_T\}$, where a so-called return-to-go (shorten as rtg) term $\Rtg_t$ is defined as $\Rtg_t \defeq \sum_{t}^T r_t$. 
This sequence serves as the input to a causal transformer that autoregressively predicts subsequent tokens. The transformer is trained to minimize the mean squared error (\mse) on action tokens during training. During evaluation (rollout), starting from a specified $\Rtg_0$ and $s_0$, for time $t$, the sequence extends with the action $a_t$, output from the transformer model, the rtg $\Rtg_{t+1} = \Rtg_0 - \sum_{0}^t r_t$ and the state $s_{t+1}$, where the reward $r_t$ and $ s_{t+1}$ are generated by the environment. The process keeps running until the episode terminates or a maximum timestep $T$ is reached. However, 
DT fails to generalize to new unseen tasks for \OMRL\ because it assumes all training data are from a single task. For example, aggregating demonstration trajectories of a robot running forward and backward as  from a single task leads to a random policy.

\paragraph{In-Context Learning (ICL) and Prompt Tuning.}\ Causal transformer architecture has demonstrated its 
superiority for solving conditional sequence generation tasks denoted as $\Prob_{\theta}(Y|X)$. Here $X$ denotes a series of input tokens, and  $Y$ is a set of output tokens. $\theta$ represents the parameters of the casual transformer. Recently, prompting-based in-context learning has enabled pretrained causal transformers to  solve a new task at inference time 
without model fine-tuning. Prompts typically include a task description and/or a few examples of this new task. ICL enables a single transformer model to solve many different tasks simultaneously. A relevant research field is prompting-tuning~\citep{lester-etal-2021-power}, which models the conditional sequence generation as $\Prob_{\{\theta_{\Prt};\theta\}}(Y | [\Prt;X])$. In the aforementioned prompt tuning formulation, the notation `$\Prt$' describes learnable soft prompt tokens (normally continuous embedding vectors) and is learned  together with parameters $\theta_{\Prt}$ via backpropagation,  while keeping the transformer's parameters, $\theta$, fixed. This work's soft token learning strategies relate to prompt tuning. 

\paragraph{Prompt Decision Transformer (PDT).} \ 
 PDT recasts the \OMRL\ challenge as a conditional sequence generation problem with in-context prompting. In DT and PDT, rtg, state, and action tokens go through three different embedding layers to project them to the same dimension. The projections are then added with their corresponding time embedding, also encoded into the same dimension with a lookup table, before feeding into the causal transformer. During the evaluation, for a new task $\mdpj$, PDT takes a trajectory prompt to guide the action generation:
 \begin{align}
 \begin{split}
  &\Prt_j  \defeq \sample(\Dsetj) \\ & =  \{[\Rtg_{j,d}^{z, *},\  s_{j,d}^{z, *},\  a_{j,d}^{z, *},\ \cdots,\ \Rtg_{j,d+m'}^{z, *}, \ s_{j,d+m'}^{z, *},\  a_{j,d+m'}^{z, *}] \},
 \label{eq:protpdt}
 \end{split}
\end{align}
where $z$ denotes the $z$-th demonstration trajectory from the set $\Dsetj$. Here $\texttt{Segment}$ denotes sampling a few time steps ($m'$) from its trajectory argument. The notation $*$ represents the segments from the demonstration $\Dsetj$. $d$ represents the starting timestep of the trajectory prompt $\sample(\Dsetj)$. This task's task-specific policy $\pipolicyj$ is derived in-context (with no gradient updates) via conditioning the learned decision transformer's autoregressive action generation on the prompt $\Prt_j$ prepended in $\Dsetoj$.

\begin{figure*}[t]
    \centering
    \includegraphics[width=0.9\textwidth]{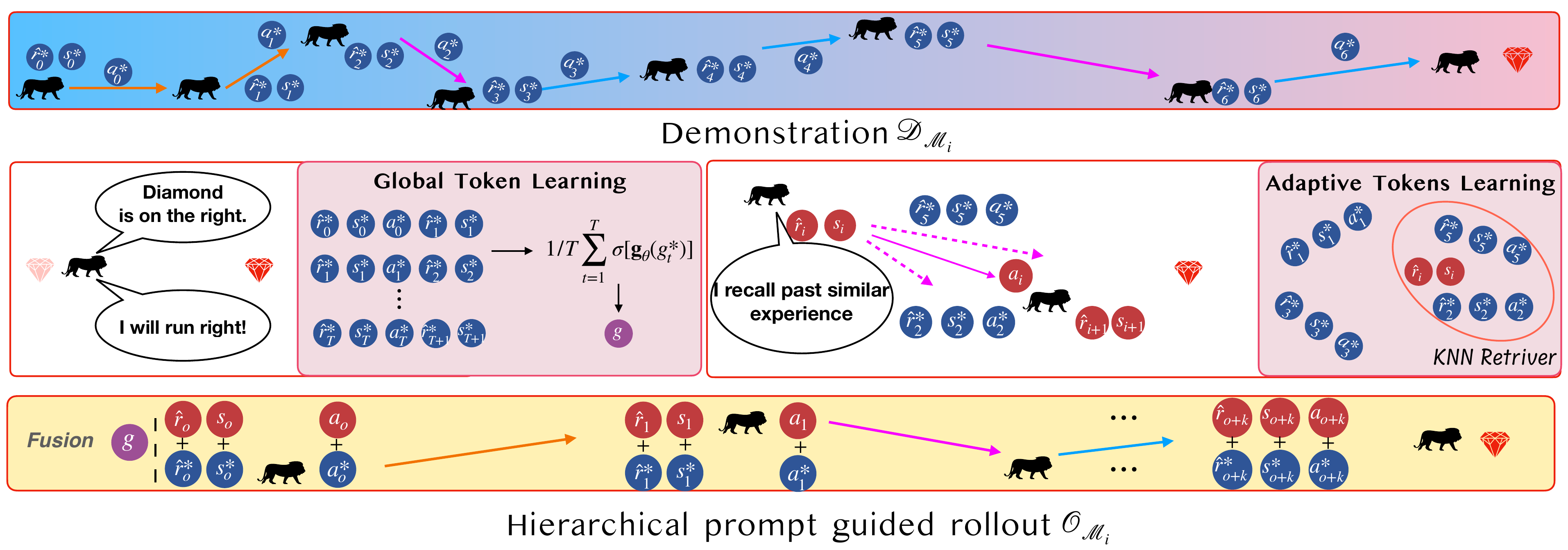}
    \vspace{-2mm}
    \caption{ The overall framework of the proposed HPDT. The top panel shows a demonstration trajectory from $\Dseti$ with a fictional blind lion looking for a red diamond. Blue circles describe different time stamps. The left part of the middle panel shows how we learn soft \tokeng to capture the transition dynamic and reward function (by using the blue circles of the demonstration). The right part of the middle panel illustrates how to retrieve \tokena for specific time stamps in rollout using \knn\ against the upper demonstration. The bottom panel shows how we fuse  the \tokeng (purple circle) and the \tokena (blue circles) within a current rollout trajectory $\Dseto$.}
  \label{fig:framework}
\end{figure*}

\section{Method}
\label{sec:method}

Recasting the \OMRL\ as a conditional sequence generation problem is promising via combining causal transformer architectures and prompting paradigm. They together enable few-shot policy generalization to new tasks with no model fine-tuning. This formulation, however, requires careful prompt design for different tasks (originates from the multi-task setup) and action generation at different timesteps (originates from the sequence data). PDT adopted a simple solution. In PDT, for each rollout segment, the prompt, which guides the action generation, is sampled with no prior and this static prompt segment provides same guidance throughout all timesteps. While simple, this heuristics-driven approach can be suboptimal for \OMRL\ when the prompt is not representative of the task. Additionally, PDT can not provide timestep-specific guidance for action generation at different $t$. Concretely, the prompt $\sample(\Dseti)$ is sampled and then fixed over the entire rollout horizon. It indicates that the action generation at different time steps $t$ will always condition on $\sample(\Dseti)$. Besides, PDT is inefficient for task recognition. $m$ snapshots of the task's transition dynamic and reward pattern to define its identity requires a prompt of length $3(m+1)$.

We analyze what are desired prompt for \OMRL. Offline RL sequences have a unique property: for each $t$ and task $\mdpi$, adjacent tokens $\hat{r}_{i,t}, s_{i,t}, a_{i,t}$ are with different modalities, and the transition from timestep $t \to t+1$: $\hat{r}_{i,t}, s_{i,t}, a_{i,t} \to \hat{r}_{i,t+1}, s_{i,t+1}$, is fully determined by the transition dynamic $\Kerneli$ and the reward function $\Rewardi$, which capture the task identity of $\mdpi$. Given the demonstration $\Dseti$, the agent should learn the task identity to guide the future action generations. For example, given two tasks that aim to teach a robot to run as fast as possible towards a pre-specified direction, forward or backward, the robot should identify the target direction from the demonstrations before moving. This global information encapsulates task-level information and is invariant within each episode. The transition dynamics and reward function are global and invariant. Local and adaptive guidance is desired for decision making at different timesteps. As in the above example, after identifying the running direction, the agent can efficiently derive the optimal action based on similar experience available in a demonstration set. Therefore, we design \textbf{hierarchical prompt} to improve the few-shot policy generalization. Given a sampled rollout trajectory segment $\sample(\Dseto)$ from $\mdpi$:
\begin{align}
     [\Rtg_{i,o}^z, \ s_{i,o}^z, \ a_{i,o}^z, \cdots, \Rtg_{i,o+m}^z, \ s_{i,o+m}^z, \ a_{i,o+m}^z],
\label{eq:rollout}
\end{align}
we randomly select a trajectory, noted as $\sample(\Dseti)$, from a handful of demonstrations. We learn a \textbf{global token} and \textbf{adaptive tokens} from $\sample(\Dseti)$. The global token captures task-level information and the adaptive tokens customize the guidance for each timestep $t$. We illustrate the overall model design in Figure~\ref{fig:framework}.

\paragraph{Learning Global Token.}\ Offline RL sequences are composed of tokens from contiguous timesteps. Each timestep includes tokens of different modalities. When generating an RL sequence, the transition dynamics and the reward function determine the transition across different timesteps. So we propose to learn the \tokeng $\conditiong$ by summarizing the RL transition dynamic and reward pattern from timesteps $\{t \to t+1\}_{t=0}^{T-1}$ in the demonstration trajectory $\sample(\Dseti)$. The global token helps the agent to distinguish different tasks. The guidance from $\conditiong$ is invariant across all timesteps for each rollout sequence.

Every data tuple $(\Rtg_{i,t}^{z, *}, s_{i,t}^{z, *}, a_{i,t}^{z, *}, s_{i,t+1}^{z, *}, \Rtg_{i,t+1}^{z, *})$ contains a snapshot of the $\Kerneli$ and $\Rewardi$. We concatenate the data tuple along the feature dimension as one vector for the global token learning. Assume $\sample(\Dseti)$ contains $T$ such transition tuples, we apply the mean aggregator as set operator to learn the global token $\conditiong$ to enjoy its permutation invariant property~\citep{zaheer2017deep,wang2022stmaml}:
{\small{
\begin{align}
\conditiong = \dfrac{1}{T}\sum_{t=0}^T \msigma(\mg( [\Rtg_{i,t}^{z, *},\  s_{i,t}^{z, *},\  a_{i,t}^{z, *}, \ s_{i,t+1}^{z, *},\  \Rtg_{i,t+1}^{z, *}])),
\label{eq:learntokeng}
\end{align}}}
where $\msigma$ is the $\gelu$ activation, $\mg$ is a linear layer with learnable parameters $\theta_g$.

The global token design enjoys the following benefits: (1). The transition dynamic and reward function are summarized into the global token $\conditiong$. Learning variables with clear physical meaning benefits meta-learning~\citep{bengio2019meta}. (2). The mean operator is length agnostic. In \method, the transition dynamic and the reward function are always summarized into one token. However, in PDT, they are conveyed in the prepended segment. $T$ glimpses of them require a prepended sequence of length $3(T+1)$; (3). The mean operator can suppress the noise variance by $1/T$ compared to other set learning approaches.

Same as all DTs, \methods uses a causal transformer~\citep{radford2019language} for autoregressive sequence modeling. The output of a token depends on its previous tokens. To guarantee the global token $\conditiong$ will guide the action generation at all timsteps $t$, we prepend it right before $\sample(\Dseto)$. The augmented sequence becomes:
\begin{align}
    [\conditiong,\ \Rtg_{i,o}^z, \ s_{i,o}^z, \ a_{i,o}^z, \cdots, \Rtg_{i,o+m}^z, \ s_{i,o+m}^z, \ a_{i,o+m}^z].
\label{eq:augseqc}
\end{align}

It ensures the action generation is guided by both transition dynamic $\Kerneli$  and reward function $\Rewardi$.

\paragraph{Learning Adaptive Tokens.}
The guidance from the global token $\conditiong$ can be insufficient for specific timesteps. To address this, we further condition the action generation on adaptive tokens that provide guidance relevant to the context.

Human recalls similar experience when making decisions. Looking back to the old experience helps adapt those relevant old solutions to meet new demands. We learn local adaptive tokens following the same spirit. At each $t$, the action $a_{i,t}^z$ heavily depends on the current rtg $\hat{r}_{i, t}^z$ and the state $s_{i, t}^z$. We look back to the demonstration trajectory by retrieving the top-relevant experience. Concretely, we compare the similarity between $[\hat{r}_{i, t}^z, s_{i, t}^z]$ with those rtg-state pairs in $\sample(\Dseti)$ and retrieve the top-$k$ similar rtg-state-action tuples:
\begin{align*}
    \{[\hat{r}_{i, t, k}^{z,*},\  s_{i, t, k}^{z,*},\  a_{i, t, k}^{z,*}]\} = \knn([\hat{r}_{i, t}^z, s_{i, t}^z]  \myarrow
    \sample(\Dseti)),
\end{align*}
where $\myarrow$ represents the Euclidean distance comparison and the retrieval process. To summarize those top-$k$ tuples, we use their mean as the final adaptive tokens at $t$. 
\begin{align}
     [\hat{r}_{i, t}^{z,*},\  s_{i, t}^{z,*},\  a_{i, t}^{z,*}] = \dfrac{1}{k}\sum_k \ma([\hat{r}_{i, t, k}^{z,*},\  s_{i, t, k}^{z,*},\  a_{i, t, k}^{z,*}]),
\label{eq:knn}
\end{align}
where $\ma$ is a linear layer with learnable parameters $\theta_a$.

Repeating the process for $t\in [o,m]$. We retrieve a template trajectory from $\sample(\Dseti)$ for the rollout trajectory in Eq.\eqref{eq:rollout}. This template sequence is:
\begin{align}
    [\Rtg_{i,o}^{z, *},\  s_{i,o}^{z, *},\  a_{i,o}^{z, *}, \cdots, \Rtg_{i,o+m}^{z, *}, \ s_{i,o+m}^{z, *},\  a_{i,o+m}^{z, *}].
\label{eq:template}
\end{align}

At each timestep $t$, the knowledge is retrieved based on only the current status (rtg-state pair in this case). Compared with the prompt sequence in PDT, see Eq.\eqref{eq:protpdt}, the retrieved template sequence in Eq.\eqref{eq:template} is customized for each $t$. Adaptive tokens complement the global token by providing context-aware guidance for action generation.

Integrating the adaptive tokens in Eq.\eqref{eq:template} into the augmented sequence in Eq.\eqref{eq:augseqc} should be training efficient. Any pending style will double the sequence length. Moreover, the multi-modality property of the offline RL sequences will limit the capacity of the causal transformer~\citep{wang2023a}, which is proposed for unimode text sequence modeling. Therefore, it is ideal that the integration will not introduce new modalities into the sequence. Therefore, we propose the summation-based adaptive tokens fusion:
\begin{align}
\begin{split}
 &[\conditiong, \ \Rtg_{i,o}^z+\Rtg_{i,o}^{z, *}, \  s_{i,o}^z + s_{i,o}^{z. *}, \  a_{i,o}^z + a_{i,o}^{z, *}, \cdots,\\ &\ \Rtg_{i,o+m}^z + \Rtg_{i,o+m}^{z, *}, \  s_{i,o+m}^z + s_{i,o+m}^{z, *}, \ a_{i,o+m}^z + a_{i,o+m}^{z, *}].
\label{eq:augseqcl}
\end{split}
\end{align}

Eq.\eqref{eq:augseqcl} satisfies the above two requirements via pairing the tokens in the rollout sequence and their corresponding tokens from the retrieved template sequence in Eq.\eqref{eq:template}.

\paragraph{Under the Hood: Learning to Embed Time Tokens.}\ 
An intelligent RL agent should also be time-aware.  The lookup table-based time encoding used in PDT is parameter-heavy and independently encodes each $t$. The parameter size of this lookup table-based embedding layer grows linearly with maximum length $T$. Also, this embedding does not consider the value and spatial relationship between time tokens when learning time representations. Both shortcomings may limit the training efficiency. There, we propose to apply \TimetoVec~\citep{kazemi2019time2vec,grigsby2021long} as a parameter-efficient mechanism for the agent to be time-aware. \TimetoVec\ projects a scalar time step $t$ to an embedding vector of $h$ dimension:
\begin{equation}
    \label{eq:t2v}
  \texttt{T2V}(t)[i]=\begin{cases}
    \omega_i t/T + \varphi_i, & \text{if~~$i=1$}. \\
    \sin{(\omega_i t/T + \varphi_i)}, & \text{if~~$1< i \leq h$}.
  \end{cases}
\end{equation} 

Here we learn parameter $\theta_t \defeq \{\omega_i, \varphi_i\}$ through backproprogation.  \TimetoVec\ contains a fixed number of parameters agnostic of the max timestep $T$. It can encode periodical events into the embedding. Moreover, adjacent timesteps are closer in the embedding space. Overall,  \TimetoVec\ is light weight, parameter efficient, and adjacency aware. 

Before feeding into the causal transformer, the global token and the retrieval-enhanced adaptive rtg, states, action tokens will go through four separate projection layers to map them to hidden spaces of the same dimension $h$ as the time embedding. The projected tokens at each timestep will be added with their corresponding time embedding vector.

\paragraph{Training \& Evaluation.}\ During training, for each task, we randomly sample a demonstration trajectory and a rollout segment. We learn both global and adaptive tokens from the demonstration, encode the time tokens with the \TimetoVec, and augment the rollout trajectory segment with the three-tier tokens. We adopt the standard teacher-forcing paradigm and train the model end-to-end to minimize the \mse\ on actions in the rollout sequence. The learnable parameters include $\theta_g, \theta_a, \theta_t$, also those in projection layers and the causal transformer.  During the evaluation, facing a new task $\mdpj$, we randomly sample a demonstration trajectory $\sample(\Dseto)$ and encode the global token $\conditionge$. With the current rtg $\hat{r}_{j, t}^z$ and state $s_{j, t}^z$, we retrieve the adaptive tokens, encode the time vectors, construct the augmented trajectory, and generate the action. The process keeps running until the episode terminates or a maximum timestep $T$ is reached. The feedback from the environment is recorded for each episode and their average across multiple episodes is used for evaluating the performance on the new task $\mdpj$. We summarize the training of \methods in Algorithm~\ref{alg:training}.

\begin{algorithm}[t]
    \caption{Hierarchical PDT Training}
    \label{alg:training}
\begin{algorithmic}
    \STATE {\bfseries Input:} training tasks $\{\mdpi\}_{i=1}^n$, causal Transformer \textit{Transformer}$_{\theta}$, training iterations $N$, rollout set $\{\Dseto\}_{i=1}^n$, demonstration set $\{\Dseti\}_{i=1}^n$, per-task batch size $M$, learning rate $\alpha$
    \FOR{$n=1$ {\bfseries to} $N$}
    \FOR{Each task $\mdp_i \in \{\mdpi\}_{i=1}^n$}
    \FOR{$m=1$ {\bfseries to} $M$}
    \STATE Sample a rollout traj {\small{$\sample(\Dseto)$}} from $\Dseto$
    \STATE Sample a demonstration traj {\small{$\sample(\Dseti)$}} from $\Dseti$
    \STATE Encode the global prompt $\conditiong$ from {\small{$\sample(\Dseti)$}}, see Eq.~\eqref{eq:learntokeng}
    \STATE Retrieve adaptive prompt for {\small{$\sample(\Dseto)$}} from {\small{$\sample(\Dseti)$}}, see Eq.~\eqref{eq:template}
    \STATE Get input $\tau^{input}_{i,m}$ by fusing the hierarchical prompts into {\small{$\sample(\Dseto)$}}, see Eq.~\eqref{eq:augseqcl}
    \ENDFOR
    \STATE Get a minibatch $ \gB_{i}^M = \{\tau^{input}_{i,m} \}_{m=1}^M$
    \ENDFOR
    \STATE Get a batch $ \gB = \{ \gB_{i}^M \}_{i=1}^N$
    \STATE $a^{pred}$ = \textit{Transformer}$_{\theta}(\tau^{input})$, $\forall \tau^{input} \in \gB$
    \STATE $\gL_{MSE} = \frac{1}{|\gB|} \sum_{\tau^{input} \in \gB}(a - a^{pred})^2$
    \STATE $\theta  \leftarrow \theta - \alpha \nabla_{\theta} \gL_{MSE}$
    \ENDFOR
\end{algorithmic}
\end{algorithm}

\section{Experiments}
\begin{table*}[t]
\caption{A summary of the evaluation environments.}
\vspace{-3mm}
\centering
\resizebox{\textwidth}{!}{
\begin{tabular}{c|c|c|c|c|c}
\toprule
\textbf{Env} & \textbf{ $\Sset \& \Aset$-dim }           & \textbf{\# Training Tasks}             & \textbf{\# Test Tasks}       & \textbf{Description} & \textbf{Variation}   \\ \hline

\CVel & 20\ \&\ 6 & 35 & 5   & A cheetah robot to run to achieve a target velocity     & Target velocity    \\
\CDir & 20\ \&\ 6 & 2 & 2   & A cheetah robot run to attain high velocity along forward or backward    & Direction   \\
\ADir & 27\ \&\ 8 & 45 & 4   & An ant agent to achieve high velocity along the specified
direction    & Goal Direction   \\
\Hopper  & 12\ \&\ 3 & 15 & 5   & An agent whose physical parameters are randomized    & Transition Dynamic   \\
\Walker  & 18\ \&\ 6 & 35 & 5   & An agent whose physical parameters are randomized  & Transition Dynamic   \\
\Reach & 39\ \&\ 4 & 15 & 5   & A Sawyer robot to reach a target position in 3D space    &  Goal Position \\
\Pick & 39\ \&\ 4 & 45 & 5   & A Sawyer robot to pick and place a puck to a goal position    &  Puck and goal positions \\
\bottomrule
\end{tabular}
}
\label{tab:envsummary}
\end{table*}

We design experiments to demonstrate the few-shot policy generalization ability of the \methods across two RL benchmarks: \DataMJ  control~\citep{todorov2012mujoco} and \DataMW~\citep{yu2020meta}. We use five continuous control meta-environments of robotic locomotion, including \CDir, \CVel, \ADir, \Hopper, and \Walker\ from \textsc{MuJoCo}. They are simulated via the \DataMJ simulator and were widely used~\citep{mitchell2021offline,rothfuss2018promp,xu2022prompting}. Concretely, \Hopper\ and \Walker\ agents differ from the transition dynamic across tasks, while the remaining three vary in terms of the reward function. Two other robotic arm manipulation environments, \Reach\ and \Pick, are from \textsc{MetaWorld}, which is designed for testing the transferability of meta-RL algorithms. See details in Table~\ref{tab:envsummary}.

\paragraph{Baselines.}\ 
We design baselines as follows:
\begin{itemize}[noitemsep,topsep=0pt]
\item MACAW~\citep{mitchell2021offline} formalized the \OMRL\ problem and is a classic actor-critic based method. It learns the task-specific value function and policy, %
modeled by neural networks and trained using MAML.

\item PDT recasts the \OMRL\ as a sequence modeling problem that gains significant improvement over approaches that are  actor-critic based. Therefore, PDT is an important baseline to compare against.

\item Another baseline is PDT-FT that first pretrains a PDT and then performs full  model fine-tuning for new tasks. 

\item One PDT variant, PTDT~\citep{hu2023prompt}. It first pretrained an \OMRL\ agent with PDT and then applied the zeroth-order optimization-based approach~\citep{tang2023zeroth} for prompt tuning. 

\end{itemize}

\paragraph{Experimental design and setup.} \
We design experiments to answer the following questions of HPDT: (1) Comparing against the baselines, can \methods enable  stronger in-context learning ability? (2) How does global token and adaptive tokens each help \methods individually? 
(3) Will \methods be robust to the value change of hyperparameters in both the demonstration and retriever?

We follow the data processing protocol as in PDT and normalize the states in the trajectories with the mean and variance in the demonstrations from the corresponding task. For global token $\conditiong$ and adaptive tokens (as in Eq.~\eqref{eq:template}) learning, we randomly sample a demonstration trajectory. The adaptive token retrieval is \knn\ based, we set $k=3$ as the default value for knowledge retrieval. We train the model for 5,000 epochs, where each epoch contains 10 updates. After training, for every test task, we let the agent interact with the environment for 50 episodes and use the averaged accumulated reward across all test tasks as the final evaluation. See other hyperparameters in Table~\ref{tab:Hyperparameters_prompt}. We also calculate the average accumulated reward in test tasks' demonstration trajectories $\Dsetj$ and noted it as $\bar{\mathcal{R}}(\Dsetj)$. It represents the quality of the policy for offline data collection.

\begin{figure*}[t]
\vspace{-4mm}
	\centering
        \subfigure[\ADir]{
		\label{fig:main1}
		\includegraphics[width=0.23\textwidth]{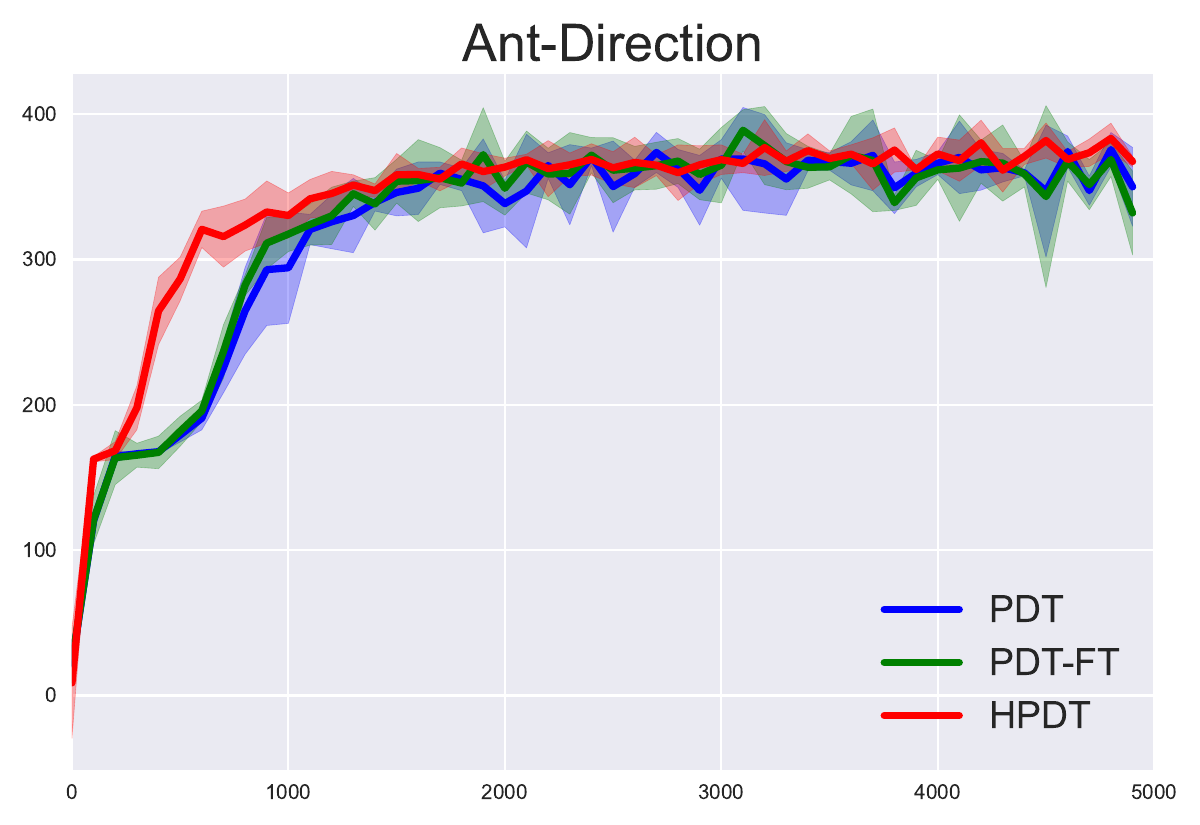}}
	\subfigure[\CVel]{
		\label{fig:main2}
		\includegraphics[width=0.23\textwidth]{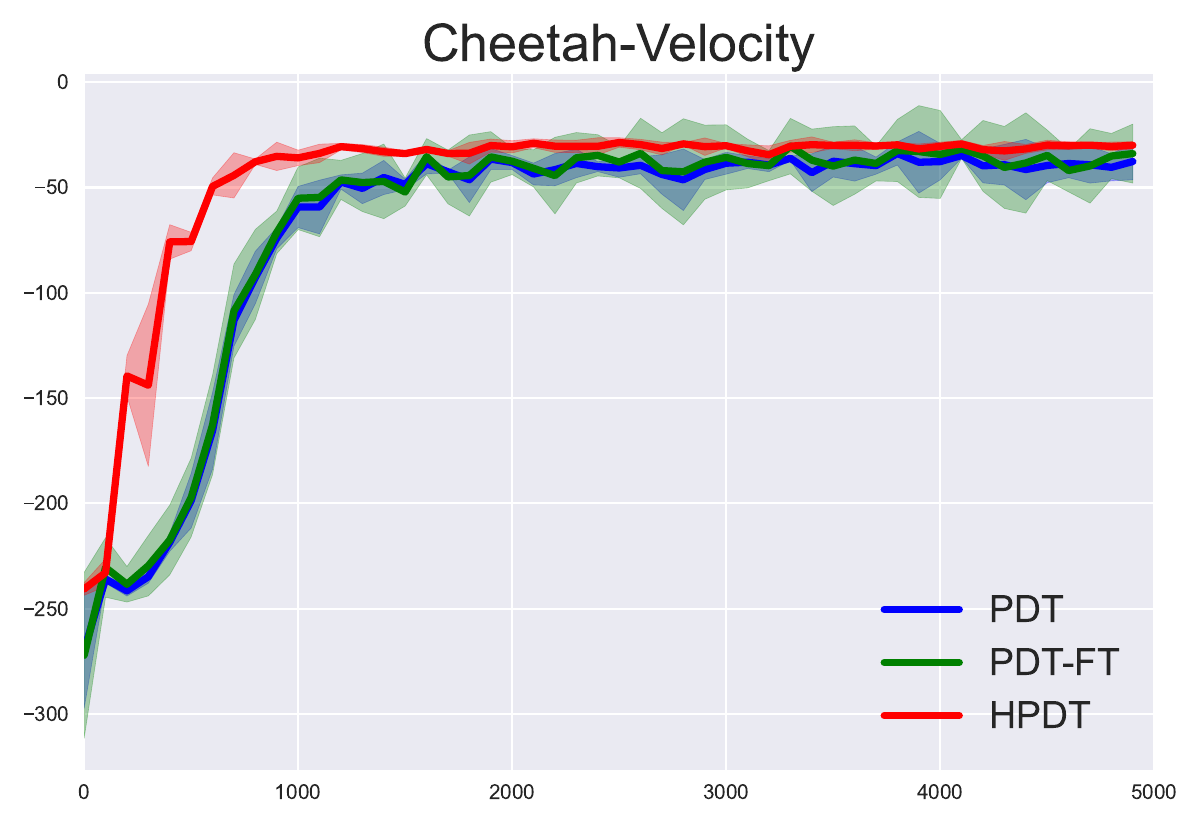}}
	\subfigure[\CDir]{
		\label{fig:main3}
		\includegraphics[width=0.23\textwidth]{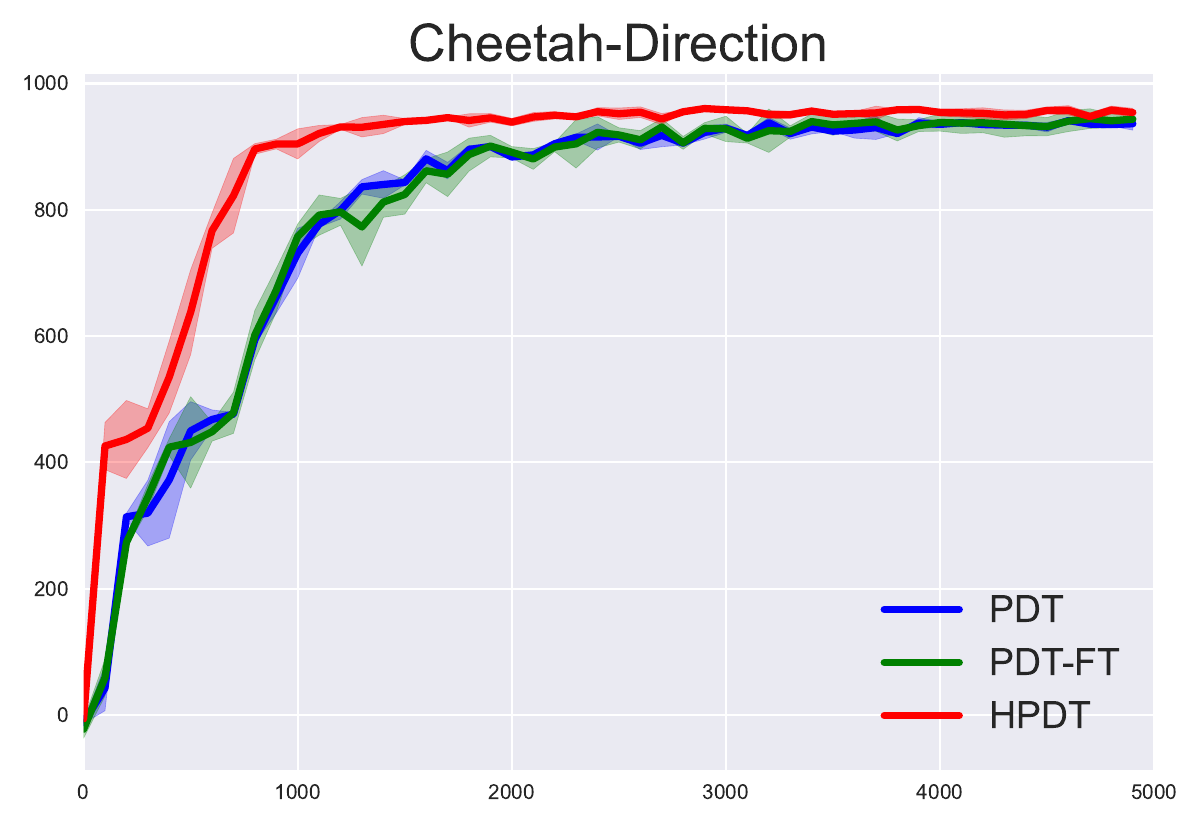}}
	\subfigure[\Pick]{
		\label{fig:main4}
		\includegraphics[width=0.23\textwidth]{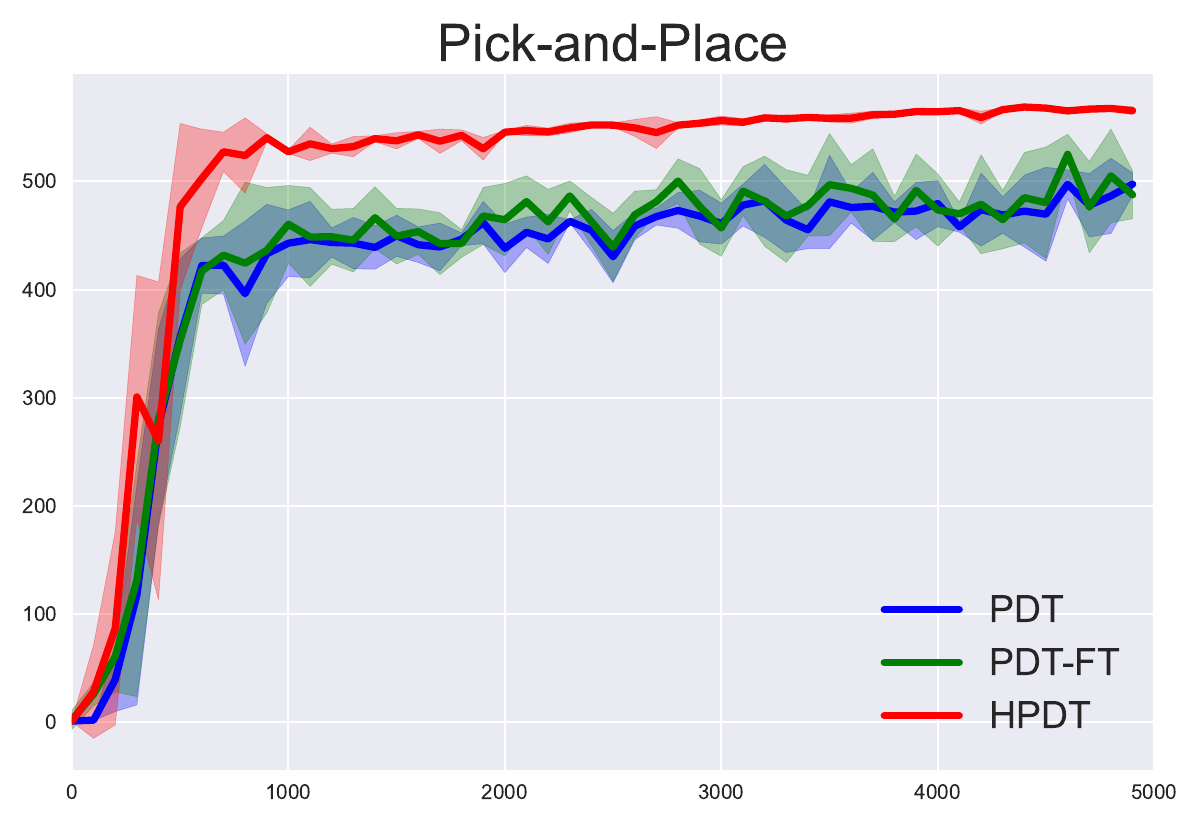}}
    \subfigure[\Reach]{
		\label{fig:main5}
		\includegraphics[width=0.23\textwidth]{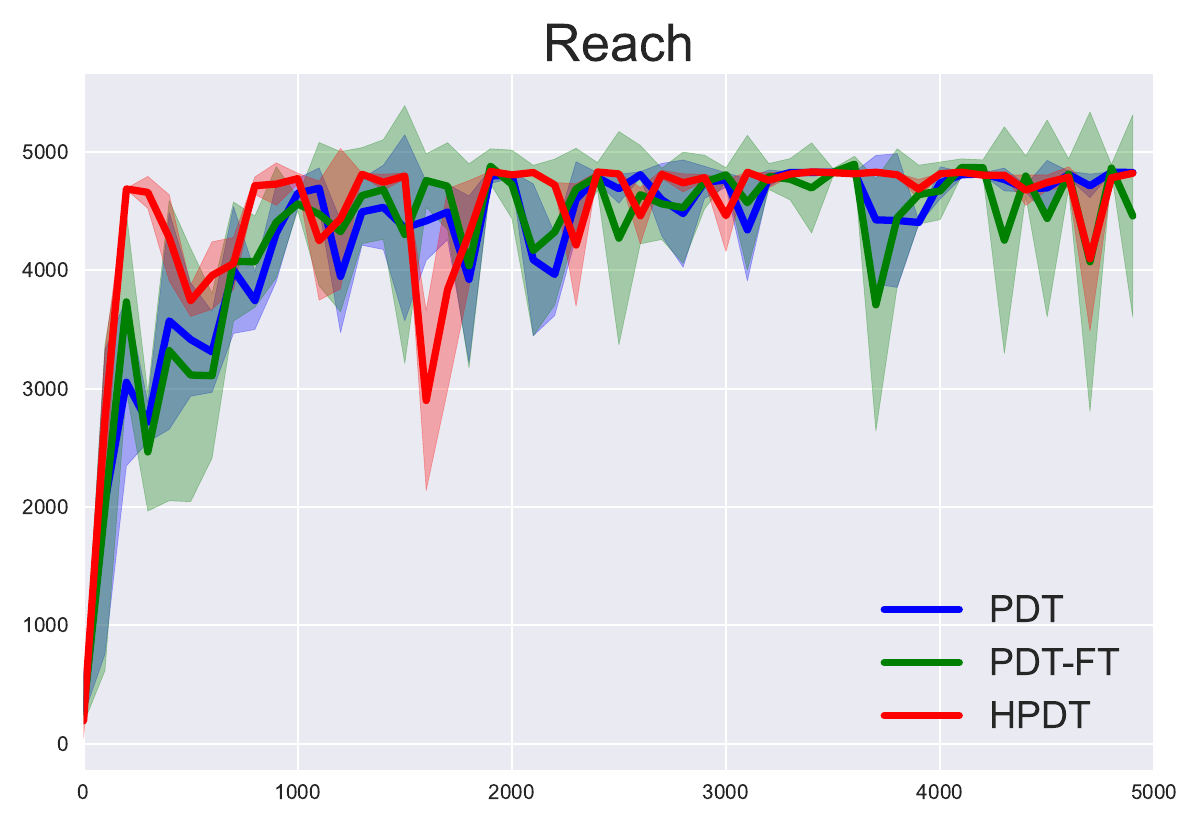}}
  \subfigure[\Walker]{
		\label{fig:main6}
		\includegraphics[width=0.23\textwidth]{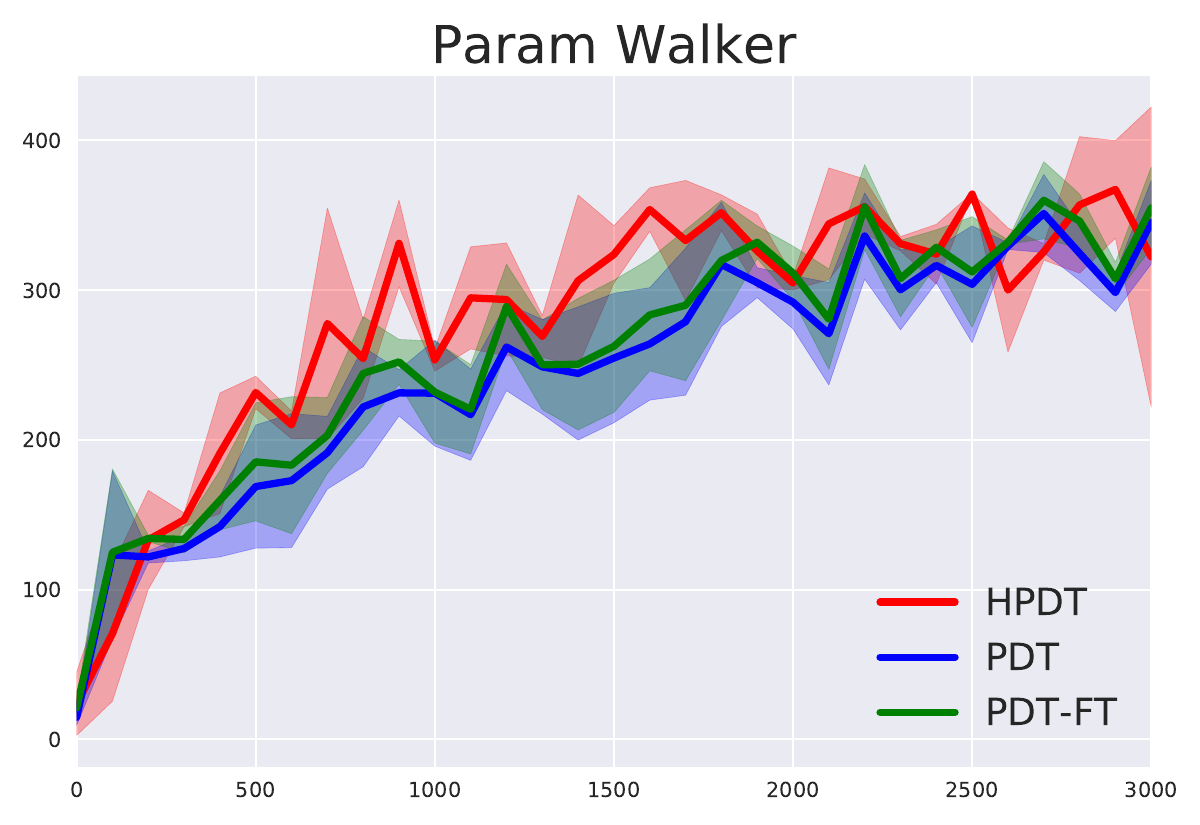}}
  \subfigure[\Hopper]{
		\label{fig:main7}
		\includegraphics[width=0.23\textwidth]{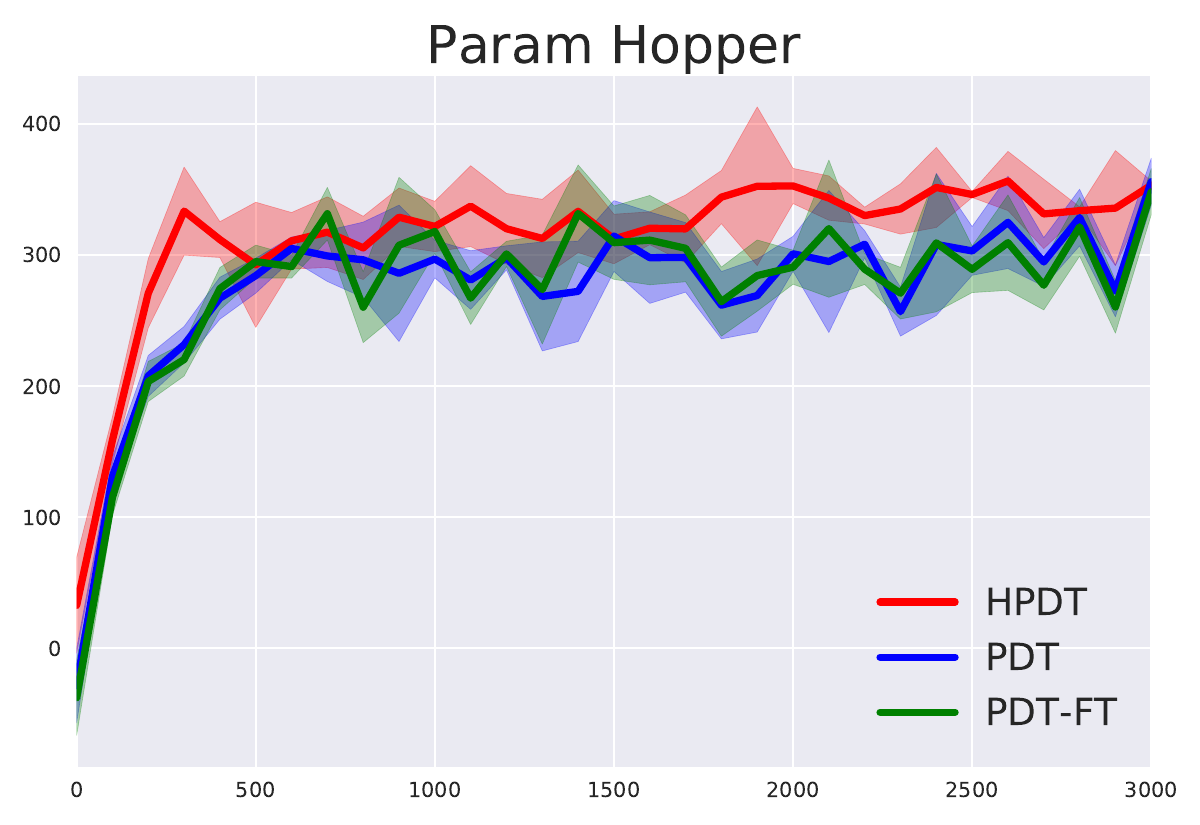}}
  \vspace{-1mm}
	\caption{Meta-testing average performance of \methods
against baselines run over three random seeds facing unseen tasks. The $x$-axis represents the training epoch and $y$-axis is the average accumulated return on testing tasks.}
	\label{fig:main}
 \vspace{-3mm}
\end{figure*}

\begin{table}[h]
    \caption{Hyperparameters of PDT, PTDT, PDT-FT, and \method. }
    \vspace{-3mm}
    \label{tab:Hyperparameters_prompt}
    \centering
    \begin{tabular}{ll}
      \toprule
      Shared Hyperparameters & Value \\
      \midrule
      $m$ (length of roll-out segment $m$)  & 20 \\
      training batch size for each task   & 16 \\
      number of evaluation episodes for each task  & 50 \\
      learning rate & 1e-4 \\
      embedding dimension & 128 \\
      \midrule
      target reward for Cheetah-dir & 1500  \\
      target reward for Cheetah-vel & 0   \\
      target reward for Ant-dir & 500   \\
      target reward for Param-Hopper & 500 \\
      target reward for Param-Walker & 450  \\
      target reward for Meta-World reach-v2 & 5000   \\
      target reward for Meta-World Pick\&Place & 650\\
      \bottomrule
    \end{tabular}
\end{table}

\paragraph{\methods achieves consistent improvements over all baselines.}\ \methods achieves the optimal results compared with MACAW, PDT, and PDT extensions including both PDT-FT and PTDT on six out of seven meta-environments.  The major results are available in Figure~\ref{fig:main} and Table~\ref{tab:results_main}. Table~\ref{tab:results_main} shows that the two PDT variations including PDT-FT and PTDT show marginal improvements over PDT, and require either extra forward or backward passes for gradient estimation. On the other hand, \methods gains significant improvements on \CVel, \CDir, \ADir, \Pick, and \Hopper\ environments. The agent trained with \methods largely surpassed the offline data collection policy $\bar{\mathcal{R}}(\Dsetj)$ on three environments and achieved closer approximations on two others including \CVel\ and \Reach. Figure~\ref{fig:main} shows that \methods is training efficient with respect to the update, especially for the training epoch 0 $\to$ 1,000. \methods quickly converges to better task-specific policies compared with other baselines.

\paragraph{\methods outperforms full fine-tuning baseline.}\
While \methods performs in-context learning, fine-tuning based approach PDT-FT updates model parameters. It calculates the gradient using the data sampled from the few-shot demonstrations of a new test task. Its extra steps of model updates may bring some benefits. However, \methods enables stronger in-context learning and provides a more efficient and adaptive strategy (last two rows of Table~\ref{tab:results_main}).

\paragraph{Global token encodes task-level information while adaptive tokens provide granular guidance.}

\begin{table}[t]
\caption{Few-shot performance of the \methods for various environments. In the table, `C-Vel', `C-Dir', and `A-Dir' represent the `\CVel', `\CDir', and `\ADir' environment respectively. We report the average and the standard deviation for three random seeds.} 
\begin{center}
\adjustbox{max width=\columnwidth}{
\begin{tabular}{l|c|c|c|c|c|c|c}
\toprule
 Models &  C-Vel  & C-Dir & A-Dir & P-Hopper & P-Walker & Reach & Pick\&Place  
 \\\midrule 
 $\bar{\mathcal{R}}(\Dsetj)$ & -23.5  & 900.4 & 351.5 & 419.9 & 438.9 & 4832.8 & 535.7 \\ \midrule
 \texttt{MACAW} & $-120.3 \pm 38.6$  & $500.8\pm 80.4$ & $253.5 \pm 3.8$ & $297.5\pm 38.7$&$328.9 \pm 43.8$ &  $3847.2\pm 74.4$ &$450.8 \pm 45.4$ \\
 \texttt{PDT} & $-37.9 \pm 4.6$  & $933.2\pm 11.4$ & $375.6 \pm 11.7$ &$328.5 \pm 21.6$ &$347.4 \pm 19.6$ & $4827.2\pm 7.3$ & $497.5 \pm 34.8$\\
 \texttt{PTDT} & $-39.5 \pm 3.7$ & $941.5 \pm 3.2$ &  $378.9 \pm 9.3$ & $ {341.6\pm12.7}$ &$ \mathbf{368.9\pm21.9}$  & $4830.5 \pm 2.9$    &$505.2 \pm 3.7$ \\ \cmidrule{1-8}

 \texttt{PDT-FT} &$-40.1 \pm 3.8$  & $936.9 \pm 4.8$ & $373.2 \pm 10.3$  & $337.2 \pm 18.3$ & $368.7 \pm 24.2$ & $4828.3\pm 6.5$ & $503.2 \pm 3.9$\\ \cmidrule{1-8} \cmidrule{1-8}
\band \methods & $\mathbf{-26.7 \pm 2.3}$ &  $\mathbf{959.4 \pm 4.0}$ & $\mathbf{383.3\pm 10.4}$&$ \mathbf{352.6\pm13.4}$ & ${367.3\pm 32.5}$ &  $\mathbf{4832.2 \pm 5.2}$  &$\mathbf{569.5 \pm 5.1}$\\
\bottomrule
\end{tabular}}
\end{center}
\label{tab:results_main}
\end{table}

\begin{figure*}[!t]
	\centering
	\subfigure[\methodap]{
		\label{fig:cd_nog}
		\includegraphics[width=0.19\textwidth]{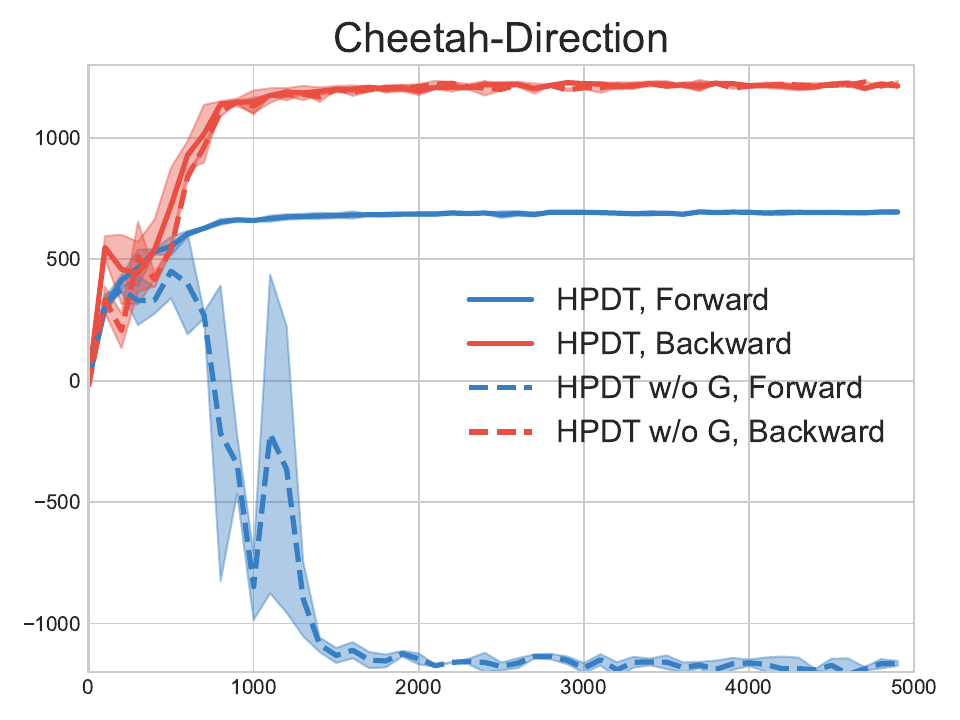}}
	\subfigure[\methodgp]{
		\label{fig:cd_noa}
		\includegraphics[width=0.19\textwidth]{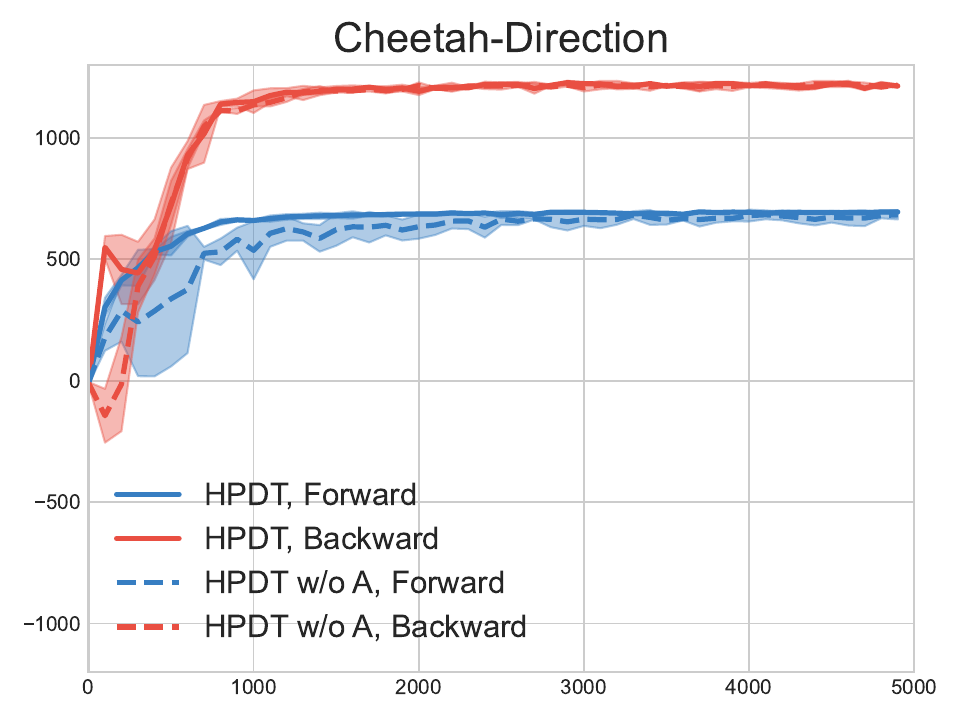}}
	\subfigure[\methodt]{
		\label{fig:cd_not}
		\includegraphics[width=0.19\textwidth]{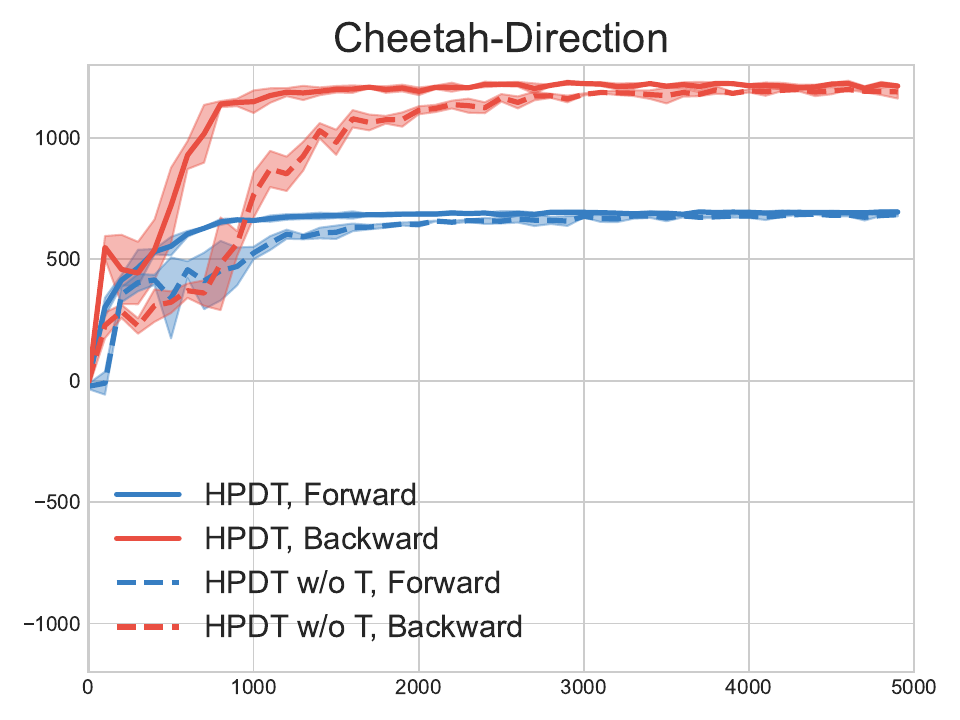}}
	\subfigure[All Variants]{
		\label{fig:cv_all}
		\includegraphics[width=0.19\textwidth]{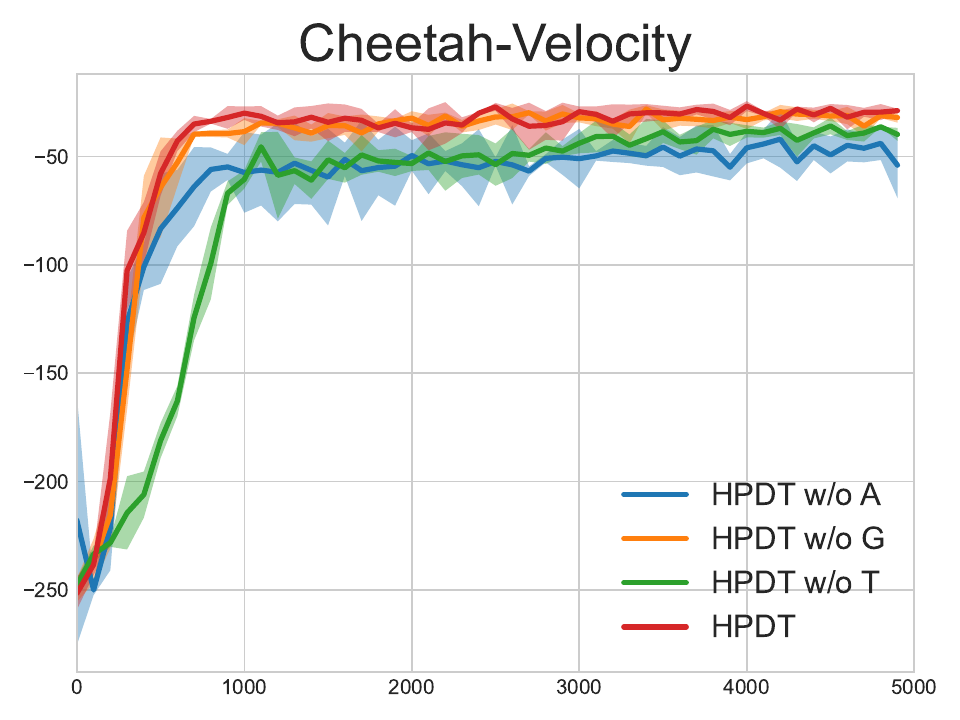}}
        \subfigure[All Variants]{
		\label{fig:pp_all}
		\includegraphics[width=0.19\textwidth]{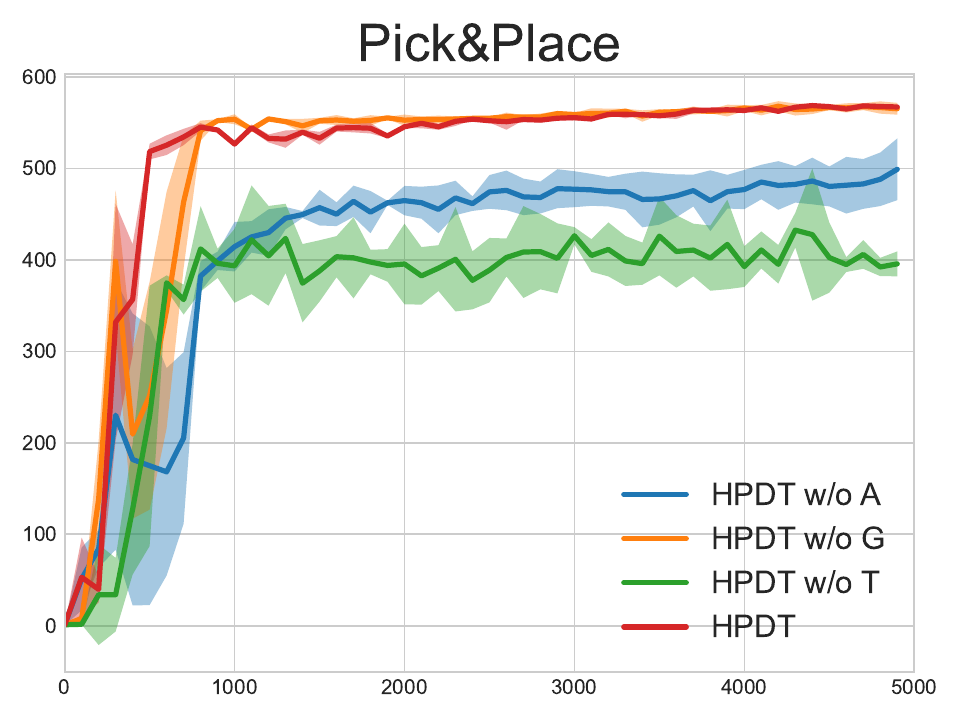}}
   \vspace{-1mm}
	\caption{ Ablation studies on \CVel ,  \CDir, and \Pick. In (a)(b)(c), we compare each ablation with the full model on \CDir. Test tasks include running forward and backward. We show the accumulated reward for each task. The solid lines represent the full model \methods for both tasks. The dashed lines represent the result of each ablation version. For \CDir, the global token is more important. In (d)(e), we show the results for \CVel\ and \Pick, where the adaptive tokens are more important. Curves represent the average accumulated reward on test tasks.}

	\label{fig:ablation}
\end{figure*}

 \ We learn two-tier tokens to guide decision transformer for \OMRL. In this subsection, we empirically investigate how each tier of token helps with ablation studies. The ablation studies are designed to isolate each component and investigate their roles. Concretely, we have two variants: \methodap, which omits global token, and \methodgp, which omits adaptive tokens. We also include \methodt, in which we replace the proposed \TimetoVec\ with the previously used lookup table. Table~\ref{tab:results_ablation} compares the results of all three variants and the full model on two robotic locomotion controls and one Sawyer robot control.

\begin{figure}[ht]
	\centering
	\subfigure[C-Vel: 35 tasks.]{
		\label{fig:proj_vel}
		\includegraphics[width=0.47\columnwidth]{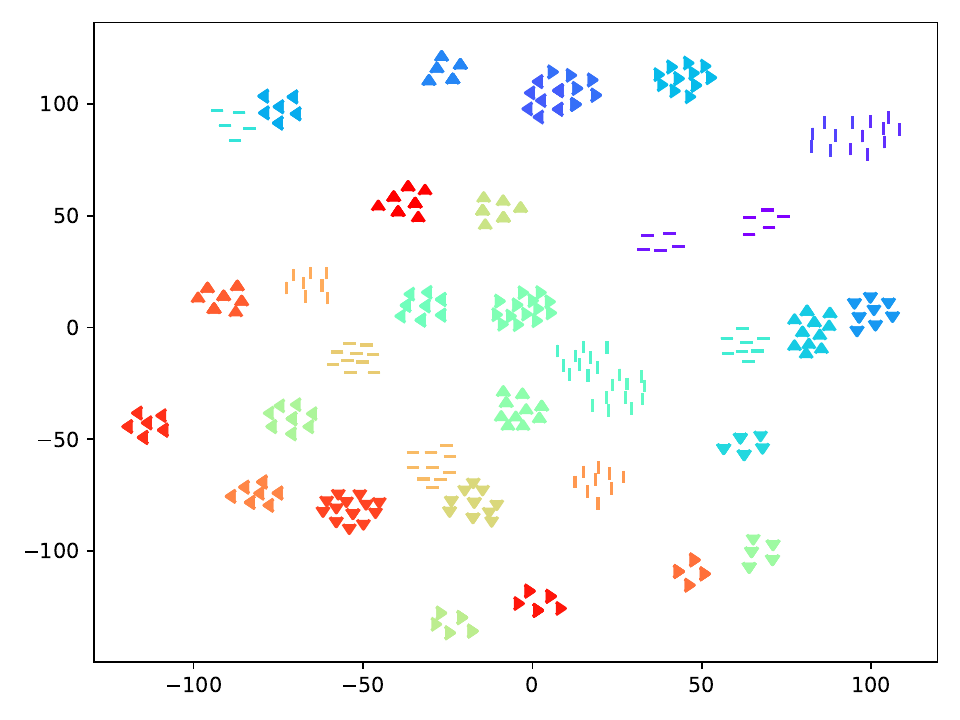}}
	\subfigure[A-Dir: 45 tasks.]{
		\label{fig:proj_ant}
		\includegraphics[width=0.47\columnwidth]{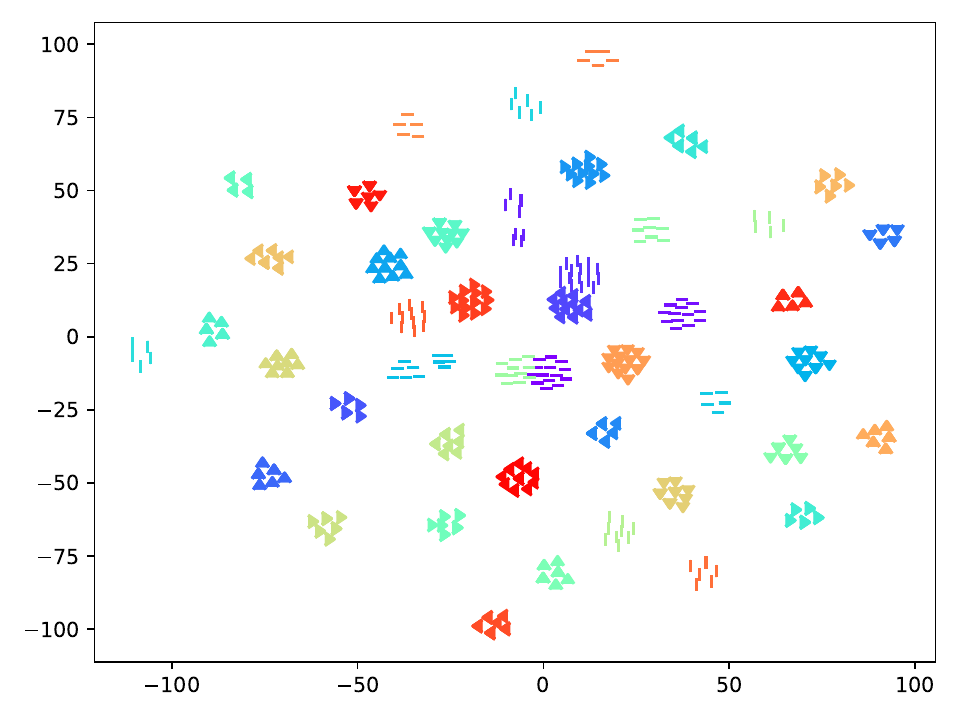}}
        \vspace{-3mm}
	\caption{2D projections of the global tokens $\conditiong$. }
\vspace{-3mm}\label{fig:projection}
\end{figure}

We design the global token $\conditiong$ to learn from the transition dynamics $\Kerneli$ and the reward function $\Rewardi$, which are necessary and sufficient conditions for task distinguishment. Without the global token $\conditiong$, the agent is confused with the task identity. For meta-learning environments where the test task identities differ significantly, the variant \methodap\ has drastically worse performance. For example, test tasks in \CDir\ include controlling a robot running to attain high velocity along either a forward or backward direction. The agent makes poor quality decisions if it fails at direction recognition. Figure~\ref{fig:cd_nog}  contains the accumulated rewards for both forward and backward test tasks. The agent trained with \methodap\ fails to recognize forward tasks, on which the accumulated reward is  $<-1,000$, see the blue dashed curve at the bottom. To further investigate the role of the $\conditiong$, we visualize their 2D projections in Figure~\ref{fig:projection}. Global tokens from different tasks $\conditiong$ are well isolated and clustered from the same task.

In some cases, task identities are more similar to each other. For example for the \CVel, as described in Table~\ref{tab:envsummary}, the variation is the target velocity, which is uniformly sampled from the range of 0 to 3. Similar task identities lead to closer task-specific policies. In this case, the role of the global token $\conditiong$ will be downplayed, and the help from adaptive tokens will be dominant. Therefore, the performance of the variant \methodgp, which removes the adaptive tokens, will be significantly impacted. See Figure~\ref{fig:cv_all}(d) for the visualizations of all variants. Similar discussion goes to the \Pick\ environment (Figure~\ref{fig:cv_all}(e)), where the goal position of the objection is uniformly sampled within a square space, see the last column in Table~\ref{tab:results_ablation}.

\begin{table}[!h]
\caption{We design ablation studies by removing the global token $\conditiong$, local tokens defined as in Eq.~\eqref{eq:template}, and replacing the \TimetoVec\ with  lookup table for time embedding.} 
\vspace{-3mm}
\begin{center}
\adjustbox{max width=0.75\columnwidth}{
\begin{tabular}{l|c|c|c}
\toprule
 Models &  C-Vel  & C-Dir & Pick\&Place
 \\\midrule 
 \cmidrule{1-4}
\methodgp  & -47.8 $\pm$ 8.1 &  950.0 $\pm$ 9.7 &  499.0 $\pm$ 33.7  \\
\methodap & -33.0 $\pm$ 1.4  & 680.8 $\pm$ 106.9 & 568.0 $\pm$ 5.5\\
\methodt & -31.3$\pm$ 4.2 &  941.6$\pm$ 4.1  &  432.5 $\pm$ 19.0\\ \cmidrule{1-4}
\band \methods & -26.7 $\pm$ 2.3 &  959.4 $\pm$ 4.0 & 569.5 $\pm$ 5.1\\
\bottomrule
\end{tabular}}
\end{center}
\label{tab:results_ablation}
\end{table}

As in Figure~\ref{fig:cd_not} and \ref{fig:cv_all}, the introduced \TimetoVec\ time embedding accelerates the convergence speed, especially at the beginning of the training phase. 
We attribute the advantage to the fewer parameters  contained in the \TimetoVec\ versus the lookup table based time embedding used by PDT.

\paragraph{\methods is robust to hyperparameter changes.}\  \methods effectively achieves \OMRL\; objectives requiring only a few rtg-state-action tuples instead of full episode demonstrations. Here we design an experiment to demonstrate \methods's performance using limited tuples for both global and adaptive token learning. We vary the tuple count ($m'$) and utilize \knn\ for adaptive token retrieval, averaging the top-$k$ similar tuples (refer to Eq.\eqref{eq:knn}), making $k$ a crucial hyperparameter. Experiments with diverse $(k, m')$ combinations, detailed in Table~\ref{tab:hyperparameter}, confirm \methods's resilience to these hyperparameter changes. Remarkably, \methods maintains comparable performance even with short demonstration trajectories, exemplified by $m'=10$, indicating its data efficiency and flexibility for handling demonstration length.

\begin{table}[H]
\centering
\caption{The robustness of \methods for different hyperparameter combinations.}
\vspace{-3mm}
\resizebox{1\columnwidth}{!}
{
\begin{tabular}{lcccccccc}    \toprule
    \multirow{3}{*}{Env} &\multicolumn{4}{c}{ $m'=10$} & &\multicolumn{3}{c}{ $m'=25$} \\
    \cmidrule{2-5}  \cmidrule{7-9}
     & {$k=3$} & {$k=5$} & {$k=7$} & {$k=10$} & & {$k=3$}  &  {{$k=5$}} &  {{$k=7$}}\\
    \midrule

    A-Dir & 369.7$\pm$16.7 & 380.0$\pm$2.4 &378.0$\pm$4.7 & 375.9$\pm$5.9  & &  374.7$\pm$  1.7&\textbf{390.5$\pm$5.3}&{382.7$\pm$4.9}\\
    C-Dir & 958.5$\pm$9.0  & 962.8$\pm$4.3 & 963.1$\pm$6.7 & 952.5$\pm$ 5.2 & & 962.4$\pm$ 7.0& \textbf{963.8 $\pm$ 3.1}&{952.7 $\pm$ 5.8}\\
    C-Vel & -28.1$\pm$4.5  & -26.5$\pm$1.4 & -29.0$\pm$ 3.7 & -30.6$\pm$ 5.7 & &  \textbf{-25.7$\pm$1.5} & -27.4$\pm$1.7& -28.7$\pm$3.2\\
    \bottomrule
\end{tabular}\label{tab:hyperparameter}}
\vspace{-3mm}
\end{table}

\paragraph{Analysis on \method's extra time cost.}

\begin{table}[H]
\caption{Time cost (PDT vs HPDT). We report the average inference time for performing \underline{five full episodes} (averaged over all testing tasks).} 
\vspace{-3mm}
\begin{center}
\resizebox{\columnwidth}{!}{
\begin{tabular}{l|c|c|c|c|c|c}
\toprule
 Models &  C-Vel  & C-Dir & A-Dir  & Pick\&Place  & P-Hopper & P-Walker
 \\\midrule 

 {PDT} & $7.1 \pm 0.6s$  & $2.9\pm 0.1s$ & $8.1 \pm 0.1s$ & $4.2\pm 0.1s$ & $4.3\pm 0.1s$ & $2.9\pm 0.5s$\\
\cmidrule{1-7}
 HPDT & ${9.0 \pm 0.3s}$ &  ${3.4 \pm 0.1s}$ & ${13.8\pm 0.2s}$ &${6.5 \pm 0.1s}$ & $5.0\pm 0.3s$  & $3.4\pm 0.8s$ \\
\bottomrule
\end{tabular}}
\end{center}
\label{tab:time_comparison}
\end{table}
Global token learning involves a concatenation operator followed by a small MLP, resulting in negligible time cost. The major time burden arises from adaptive token learning.

During training, the computational cost remains insignificant due to matrix operations. By having all $\hat{r}_t, s_t$ and $\hat{r}_t^*, s_t^*$ for $t\in [0, T]$ available, we can simultaneously compute their similarity for all $t$ using matrix operations. We simplify this process by comparing the distances of two matrices $X\in \mathbf{R}^{bs, l, d}$ and $Y\in \mathbf{R}^{bs, T, d}$, where $bs$ represents batch size, $l$ denotes sequence length, and $d$ is the embedding dimension. The distance matrix $D \in \mathbf{R}^{bs, l, T}$ is computed, with each element $D_{k, i, j}$ representing the $L_2$ distance between $X_{k, i}$ and $Y_{k,j}$, utilizing the efficient function $torch.cdist$.

During inference, the state transits from  $s_t$ to $s_{t+1}$ via real-time interaction with the environment. Therefore, distance comparison is needed for each $t$.  
In  Table~\ref{tab:time_comparison}, we empirically compare the time cost of performing 5 full episodes (trajectories) averaged over all testing tasks. Indeed, HPDT has a higher inference time cost than PDT. However, a long-running episode is preferred for tasks like A-Dir and C-Vel. Part of the time duration increase (PDT vs HPDT) is attributed to that HPDT-based RL agents learn a better policy than PDT (therefore running longer episodes).

\section{Related Work}

\textbf{Offline RL as Sequence Generation.}\ Treating policy learning in offline RL as a sequence generation problem via the language model is gaining momentum since DT~\citep{chen2021decision}. Concurrent work is trajectory transformer~\citep{janner2021sequence}. TT discretizes independently every dimension of the state, action, and reward. It models both the environment and policy. During the evaluation, TT adapts beam search for planning. Bootstrapped Transformer~\citep{wang2022bootstrapped} incorporates the idea of bootstrapping and leverages the learned model to self-generate more offline data to further boost the sequence model training. ESPER~\citep{paster2022you} analyzes that DT fails in the stochastic environment because the rtg term depends on environment stochasticity. It proposes to cluster trajectories and conditions the learning on average cluster returns. Brandfonbrener~\etal\cite{brandfonbrener2022does} theoretically show that the successful scenarios for the return-conditioned decision transformer would require a stronger assumption on the sample complexity. Furuta~\etal\cite{furuta2021generalized} suggests that DT is performing hindsight information matching and generalizing DT by replacing the rtg term with other statistics of the future trajectory. 

\textbf{\OMRL.}\ Offline meta reinforcement learning (\OMRL) targets approximating the task-specific optimal policy given a handful of static demonstrations from the task. MACAW~\citep{mitchell2021offline} formalizes the \OMRL\ setup and proposes to combine MAML with value-based RL. It increases the expressive power of the meta-learner by using the advantage regression as a subroutine in the inner loop. 
A popular line of \OMRL\ methods adapt existing online meta-RL approaches, which rely on context-conditioned policy trained by TD-learning~\citep{sutton2018reinforcement}. They may overestimate or underestimate the reward function, and finally lead to suboptimal performance.

To the best of our knowledge, PDT~\citep{xu2022prompting} is the first work reframing the \OMRL\ as a conditional sequence generation problem. It gains significant improvements over previous \OMRL\ methods by leveraging the transformer architecture's strong ability to learn from a few examples and then generalize. Both \methods and PDT distill the policy in the offline dataset to the DT. AD~\citep{laskin2022context} proposes to distill the RL algorithm to the DT by collecting a large enough offline dataset covering the learning history of the algorithm. MetaDiffuser~\citep{ni2023metadiffuser} adopted the double-phase training strategy, which first trains a task-oriented context encoder, and then  trajectories can be labeled with the trained encoder to perform conditional generation with diffusion model.
Another loosely-related line of work is under the multi-task RL angle~\citep{reed2022generalist,goyal2022retrieval,furuta2021generalized,sodhani2021multi,yu2020gradient,zhang2020learning}. However, the main target of the multi-task RL is learning one agent to handle all training tasks, instead of generalizing to future unseen tasks.

\textbf{Hierarchical RL.}\ Hierarchical approaches are also being applied to the RL domain~\citep{barto2003recent,hengst2010hierarchical,nachum2018data,vezhnevets2017feudal,10342230}. They usually comprise a low-level controller and a high-level planner. A high-level planner learns to select optimal subtasks as the higher-level actions, and each subtask itself is a RL problem solved by the low-level controller. In \method, the hierarchy arises from the two-tier knowledge learned from the demonstration set. Lastly, the idea of using the retrieved knowledge from past experience, episodic memory, or the replay buffer to aid decision-making has also been  studied~\citep{zhu2019episodic,eysenbach2019search,lewis2020retrieval,goyal2022retrieval}. However, they haven't been studied in the \OMRL\ setting.

\textbf{Retrieval-Enhanced Transformers.}\ Rarely developed for RL, retrieval-enhanced transformers for NLP are well-explored. In the NLP domain, a small language model equipped with a retrieval module is capable of achieving on-par performance on various tasks compared with large language models~\citep{borgeaud2022improving,khandelwal2019generalization,izacard2020leveraging}. The pretrained language models save the world knowledge in parameters and the retrievers capture the factual knowledge in a modular and interpretable paradigm. REALM~\citep{guu2020retrieval} firstly proposes to jointly train end-to-end a retrieval system with an encoder language model for open-domain QA. Atlas~\citep{izacard2022few} trained a retriever together with a seq2seq model and demonstrated its strong few-shot learning capabilities on various language tasks. It outperforms a 540B parameters model despite having 50x fewer parameters. RAG~\citep{lewis2020retrieval} designs finetuning approach for language models and neural retrievers for language generation. Besides, Peng~\etal\citep{peng2019text} retrieve
exemplar text from training data as `soft templates' for text summarization; Li~\etal\cite{li2018transfer} design lexical-level similarity based retrieval for text style transfer; UVLP~\citep{zhou2022unsupervised} propose retrieval-based multi-granular alignment for vision-and-language cross-modality alignment, etc.

Recent literature includes a growing body of LLM orchestration \citep{yao2022react, shen2024hugginggpt} studies. These methods need extensive compute resources and are costly. DTs are complimentary and offer a more lightweight and focused approach to sequential decision-making, specifically optimized for learning from behavioral data without requiring the broad language understanding capabilities of full LLMs.

\vspace{-3mm}
\section{Conclusions}
In this paper, we introduce HPDT, a method designed for Offline Meta Reinforcement Learning that utilizes hierarchical prompting to effectively leverage structural information present in demonstrations. This method employs global and local adaptive prompts derived from few-shot demonstration sets to guide action generation during rollouts. We start by learning global tokens to provide task-level guidance related to transition dynamics and reward function of a new task. We, then, learn adaptive tokens by retrieving relevant prompt segments from demonstration trajectories. These two levels of tokens guide the Decision Transformer in generating a sequence of actions for new, previously unseen RL tasks. \methods outperforms existing state-of-the-art prompting-based baselines by providing the Decision Transformer with more targeted and contextual guidance. Furthermore, our method is robust to variations in hyperparameters. The results suggest that hierarchical prompting is an effective strategy for few-shot policy generalization based on decision transformers.

\bibliographystyle{ACM-Reference-Format}
\bibliography{sample-base}


\begin{thebibliography}{55}


\ifx \showCODEN    \undefined \def \showCODEN     #1{\unskip}     \fi
\ifx \showDOI      \undefined \def \showDOI       #1{#1}\fi
\ifx \showISBNx    \undefined \def \showISBNx     #1{\unskip}     \fi
\ifx \showISBNxiii \undefined \def \showISBNxiii  #1{\unskip}     \fi
\ifx \showISSN     \undefined \def \showISSN      #1{\unskip}     \fi
\ifx \showLCCN     \undefined \def \showLCCN      #1{\unskip}     \fi
\ifx \shownote     \undefined \def \shownote      #1{#1}          \fi
\ifx \showarticletitle \undefined \def \showarticletitle #1{#1}   \fi
\ifx \showURL      \undefined \def \showURL       {\relax}        \fi
\providecommand\bibfield[2]{#2}
\providecommand\bibinfo[2]{#2}
\providecommand\natexlab[1]{#1}
\providecommand\showeprint[2][]{arXiv:#2}

\bibitem[Barto and Mahadevan(2003)]%
        {barto2003recent}
\bibfield{author}{\bibinfo{person}{Andrew~G Barto} {and}
  \bibinfo{person}{Sridhar Mahadevan}.} \bibinfo{year}{2003}\natexlab{}.
\newblock \showarticletitle{Recent advances in hierarchical reinforcement
  learning}.
\newblock \bibinfo{journal}{\emph{Discrete event dynamic systems}}
  \bibinfo{volume}{13}, \bibinfo{number}{1-2} (\bibinfo{year}{2003}),
  \bibinfo{pages}{41--77}.
\newblock


\bibitem[Bengio et~al\mbox{.}(2019)]%
        {bengio2019meta}
\bibfield{author}{\bibinfo{person}{Yoshua Bengio}, \bibinfo{person}{Tristan
  Deleu}, \bibinfo{person}{Nasim Rahaman}, \bibinfo{person}{Rosemary Ke},
  \bibinfo{person}{S{\'e}bastien Lachapelle}, \bibinfo{person}{Olexa Bilaniuk},
  \bibinfo{person}{Anirudh Goyal}, {and} \bibinfo{person}{Christopher Pal}.}
  \bibinfo{year}{2019}\natexlab{}.
\newblock \showarticletitle{A meta-transfer objective for learning to
  disentangle causal mechanisms}.
\newblock \bibinfo{journal}{\emph{arXiv preprint arXiv:1901.10912}}
  (\bibinfo{year}{2019}).
\newblock


\bibitem[Borgeaud et~al\mbox{.}(2022)]%
        {borgeaud2022improving}
\bibfield{author}{\bibinfo{person}{Sebastian Borgeaud}, \bibinfo{person}{Arthur
  Mensch}, \bibinfo{person}{Jordan Hoffmann}, \bibinfo{person}{Trevor Cai},
  \bibinfo{person}{Eliza Rutherford}, \bibinfo{person}{Katie Millican},
  \bibinfo{person}{George~Bm Van Den~Driessche}, \bibinfo{person}{Jean-Baptiste
  Lespiau}, \bibinfo{person}{Bogdan Damoc}, \bibinfo{person}{Aidan Clark},
  {et~al\mbox{.}}} \bibinfo{year}{2022}\natexlab{}.
\newblock \showarticletitle{Improving language models by retrieving from
  trillions of tokens}. In \bibinfo{booktitle}{\emph{ICML}}. PMLR,
  \bibinfo{pages}{2206--2240}.
\newblock


\bibitem[Brandfonbrener et~al\mbox{.}(2022)]%
        {brandfonbrener2022does}
\bibfield{author}{\bibinfo{person}{David Brandfonbrener},
  \bibinfo{person}{Alberto Bietti}, \bibinfo{person}{Jacob Buckman},
  \bibinfo{person}{Romain Laroche}, {and} \bibinfo{person}{Joan Bruna}.}
  \bibinfo{year}{2022}\natexlab{}.
\newblock \showarticletitle{When does return-conditioned supervised learning
  work for offline reinforcement learning?}
\newblock \bibinfo{journal}{\emph{NeurIPS}}  \bibinfo{volume}{35}
  (\bibinfo{year}{2022}), \bibinfo{pages}{1542--1553}.
\newblock


\bibitem[Chen et~al\mbox{.}(2021)]%
        {chen2021decision}
\bibfield{author}{\bibinfo{person}{Lili Chen}, \bibinfo{person}{Kevin Lu},
  \bibinfo{person}{Aravind Rajeswaran}, \bibinfo{person}{Kimin Lee},
  \bibinfo{person}{Aditya Grover}, \bibinfo{person}{Misha Laskin},
  \bibinfo{person}{Pieter Abbeel}, \bibinfo{person}{Aravind Srinivas}, {and}
  \bibinfo{person}{Igor Mordatch}.} \bibinfo{year}{2021}\natexlab{}.
\newblock \showarticletitle{Decision transformer: Reinforcement learning via
  sequence modeling}.
\newblock \bibinfo{journal}{\emph{NeurIPS}}  \bibinfo{volume}{34}
  (\bibinfo{year}{2021}), \bibinfo{pages}{15084--15097}.
\newblock


\bibitem[Correia and Alexandre(2023)]%
        {10342230}
\bibfield{author}{\bibinfo{person}{André Correia} {and}
  \bibinfo{person}{Luis~A. Alexandre}.} \bibinfo{year}{2023}\natexlab{}.
\newblock \showarticletitle{Hierarchical Decision Transformer}. In
  \bibinfo{booktitle}{\emph{2023 IEEE/RSJ IROS}}. \bibinfo{pages}{1661--1666}.
\newblock
\urldef\tempurl%
\url{https://doi.org/10.1109/IROS55552.2023.10342230}
\showDOI{\tempurl}


\bibitem[Eysenbach et~al\mbox{.}(2019)]%
        {eysenbach2019search}
\bibfield{author}{\bibinfo{person}{Ben Eysenbach}, \bibinfo{person}{Russ~R
  Salakhutdinov}, {and} \bibinfo{person}{Sergey Levine}.}
  \bibinfo{year}{2019}\natexlab{}.
\newblock \showarticletitle{Search on the replay buffer: Bridging planning and
  reinforcement learning}.
\newblock \bibinfo{journal}{\emph{NeurIPS}}  \bibinfo{volume}{32}
  (\bibinfo{year}{2019}).
\newblock


\bibitem[Furuta et~al\mbox{.}(2021)]%
        {furuta2021generalized}
\bibfield{author}{\bibinfo{person}{Hiroki Furuta}, \bibinfo{person}{Yutaka
  Matsuo}, {and} \bibinfo{person}{Shixiang~Shane Gu}.}
  \bibinfo{year}{2021}\natexlab{}.
\newblock \showarticletitle{Generalized decision transformer for offline
  hindsight information matching}.
\newblock \bibinfo{journal}{\emph{arXiv preprint arXiv:2111.10364}}
  (\bibinfo{year}{2021}).
\newblock


\bibitem[Goyal et~al\mbox{.}(2022)]%
        {goyal2022retrieval}
\bibfield{author}{\bibinfo{person}{Anirudh Goyal}, \bibinfo{person}{Abram
  Friesen}, \bibinfo{person}{Andrea Banino}, \bibinfo{person}{Theophane Weber},
  \bibinfo{person}{Nan~Rosemary Ke}, \bibinfo{person}{Adria~Puigdomenech
  Badia}, \bibinfo{person}{Arthur Guez}, \bibinfo{person}{Mehdi Mirza},
  \bibinfo{person}{Peter~C Humphreys}, \bibinfo{person}{Ksenia Konyushova},
  {et~al\mbox{.}}} \bibinfo{year}{2022}\natexlab{}.
\newblock \showarticletitle{Retrieval-augmented reinforcement learning}. In
  \bibinfo{booktitle}{\emph{ICML}}. PMLR, \bibinfo{pages}{7740--7765}.
\newblock


\bibitem[Grigsby et~al\mbox{.}(2021)]%
        {grigsby2021long}
\bibfield{author}{\bibinfo{person}{Jake Grigsby}, \bibinfo{person}{Zhe Wang},
  \bibinfo{person}{Nam Nguyen}, {and} \bibinfo{person}{Yanjun Qi}.}
  \bibinfo{year}{2021}\natexlab{}.
\newblock \showarticletitle{Long-range transformers for dynamic spatiotemporal
  forecasting}.
\newblock \bibinfo{journal}{\emph{arXiv preprint arXiv:2109.12218}}
  (\bibinfo{year}{2021}).
\newblock


\bibitem[Guu et~al\mbox{.}(2020)]%
        {guu2020retrieval}
\bibfield{author}{\bibinfo{person}{Kelvin Guu}, \bibinfo{person}{Kenton Lee},
  \bibinfo{person}{Zora Tung}, \bibinfo{person}{Panupong Pasupat}, {and}
  \bibinfo{person}{Mingwei Chang}.} \bibinfo{year}{2020}\natexlab{}.
\newblock \showarticletitle{Retrieval augmented language model pre-training}.
  In \bibinfo{booktitle}{\emph{ICML}}. PMLR, \bibinfo{pages}{3929--3938}.
\newblock


\bibitem[Hengst(2010)]%
        {hengst2010hierarchical}
\bibfield{author}{\bibinfo{person}{Bernhard Hengst}.}
  \bibinfo{year}{2010}\natexlab{}.
\newblock \bibinfo{title}{Hierarchical Reinforcement Learning.}
\newblock
\newblock


\bibitem[Hu et~al\mbox{.}(2023)]%
        {hu2023prompt}
\bibfield{author}{\bibinfo{person}{Shengchao Hu}, \bibinfo{person}{Li Shen},
  \bibinfo{person}{Ya Zhang}, {and} \bibinfo{person}{Dacheng Tao}.}
  \bibinfo{year}{2023}\natexlab{}.
\newblock \showarticletitle{Prompt-Tuning Decision Transformer with Preference
  Ranking}.
\newblock \bibinfo{journal}{\emph{arXiv preprint arXiv:2305.09648}}
  (\bibinfo{year}{2023}).
\newblock


\bibitem[Izacard and Grave(2020)]%
        {izacard2020leveraging}
\bibfield{author}{\bibinfo{person}{Gautier Izacard} {and}
  \bibinfo{person}{Edouard Grave}.} \bibinfo{year}{2020}\natexlab{}.
\newblock \showarticletitle{Leveraging passage retrieval with generative models
  for open domain question answering}.
\newblock \bibinfo{journal}{\emph{arXiv preprint arXiv:2007.01282}}
  (\bibinfo{year}{2020}).
\newblock


\bibitem[Izacard et~al\mbox{.}(2022)]%
        {izacard2022few}
\bibfield{author}{\bibinfo{person}{Gautier Izacard}, \bibinfo{person}{Patrick
  Lewis}, \bibinfo{person}{Maria Lomeli}, \bibinfo{person}{Lucas Hosseini},
  \bibinfo{person}{Fabio Petroni}, \bibinfo{person}{Timo Schick},
  \bibinfo{person}{Jane Dwivedi-Yu}, \bibinfo{person}{Armand Joulin},
  \bibinfo{person}{Sebastian Riedel}, {and} \bibinfo{person}{Edouard Grave}.}
  \bibinfo{year}{2022}\natexlab{}.
\newblock \showarticletitle{Few-shot learning with retrieval augmented language
  models}.
\newblock \bibinfo{journal}{\emph{arXiv preprint arXiv:2208.03299}}
  (\bibinfo{year}{2022}).
\newblock


\bibitem[Janner et~al\mbox{.}(2021)]%
        {janner2021sequence}
\bibfield{author}{\bibinfo{person}{Michael Janner}, \bibinfo{person}{Qiyang
  Li}, {and} \bibinfo{person}{Sergey Levine}.} \bibinfo{year}{2021}\natexlab{}.
\newblock \showarticletitle{Offline Reinforcement Learning as One Big Sequence
  Modeling Problem}. In \bibinfo{booktitle}{\emph{NeurIPS}}.
\newblock


\bibitem[Kahneman(2011)]%
        {kahneman2011thinking}
\bibfield{author}{\bibinfo{person}{Daniel Kahneman}.}
  \bibinfo{year}{2011}\natexlab{}.
\newblock \bibinfo{booktitle}{\emph{Thinking, fast and slow}}.
\newblock \bibinfo{publisher}{macmillan}.
\newblock


\bibitem[Kazemi et~al\mbox{.}(2019)]%
        {kazemi2019time2vec}
\bibfield{author}{\bibinfo{person}{Seyed~Mehran Kazemi},
  \bibinfo{person}{Rishab Goel}, \bibinfo{person}{Sepehr Eghbali},
  \bibinfo{person}{Janahan Ramanan}, \bibinfo{person}{Jaspreet Sahota},
  \bibinfo{person}{Sanjay Thakur}, \bibinfo{person}{Stella Wu},
  \bibinfo{person}{Cathal Smyth}, \bibinfo{person}{Pascal Poupart}, {and}
  \bibinfo{person}{Marcus Brubaker}.} \bibinfo{year}{2019}\natexlab{}.
\newblock \showarticletitle{Time2vec: Learning a vector representation of
  time}.
\newblock \bibinfo{journal}{\emph{arXiv preprint arXiv:1907.05321}}
  (\bibinfo{year}{2019}).
\newblock


\bibitem[Khandelwal et~al\mbox{.}(2019)]%
        {khandelwal2019generalization}
\bibfield{author}{\bibinfo{person}{Urvashi Khandelwal}, \bibinfo{person}{Omer
  Levy}, \bibinfo{person}{Dan Jurafsky}, \bibinfo{person}{Luke Zettlemoyer},
  {and} \bibinfo{person}{Mike Lewis}.} \bibinfo{year}{2019}\natexlab{}.
\newblock \showarticletitle{Generalization through memorization: Nearest
  neighbor language models}.
\newblock \bibinfo{journal}{\emph{arXiv preprint arXiv:1911.00172}}
  (\bibinfo{year}{2019}).
\newblock


\bibitem[Kumar et~al\mbox{.}(2022)]%
        {kumar2022should}
\bibfield{author}{\bibinfo{person}{Aviral Kumar}, \bibinfo{person}{Joey Hong},
  \bibinfo{person}{Anikait Singh}, {and} \bibinfo{person}{Sergey Levine}.}
  \bibinfo{year}{2022}\natexlab{}.
\newblock \showarticletitle{When should we prefer offline reinforcement
  learning over behavioral cloning?}
\newblock \bibinfo{journal}{\emph{arXiv preprint arXiv:2204.05618}}
  (\bibinfo{year}{2022}).
\newblock


\bibitem[Laskin et~al\mbox{.}(2022)]%
        {laskin2022context}
\bibfield{author}{\bibinfo{person}{Michael Laskin}, \bibinfo{person}{Luyu
  Wang}, \bibinfo{person}{Junhyuk Oh}, \bibinfo{person}{Emilio Parisotto},
  \bibinfo{person}{Stephen Spencer}, \bibinfo{person}{Richie Steigerwald},
  \bibinfo{person}{DJ Strouse}, \bibinfo{person}{Steven Hansen},
  \bibinfo{person}{Angelos Filos}, \bibinfo{person}{Ethan Brooks},
  {et~al\mbox{.}}} \bibinfo{year}{2022}\natexlab{}.
\newblock \showarticletitle{In-context reinforcement learning with algorithm
  distillation}.
\newblock \bibinfo{journal}{\emph{arXiv preprint arXiv:2210.14215}}
  (\bibinfo{year}{2022}).
\newblock


\bibitem[Lester et~al\mbox{.}(2021)]%
        {lester-etal-2021-power}
\bibfield{author}{\bibinfo{person}{Brian Lester}, \bibinfo{person}{Rami
  Al-Rfou}, {and} \bibinfo{person}{Noah Constant}.}
  \bibinfo{year}{2021}\natexlab{}.
\newblock \showarticletitle{The Power of Scale for Parameter-Efficient Prompt
  Tuning}. In \bibinfo{booktitle}{\emph{EMNLP}},
  \bibfield{editor}{\bibinfo{person}{Marie-Francine Moens},
  \bibinfo{person}{Xuanjing Huang}, \bibinfo{person}{Lucia Specia}, {and}
  \bibinfo{person}{Scott Wen-tau Yih}} (Eds.). \bibinfo{publisher}{ACL},
  \bibinfo{address}{Online and Punta Cana, Dominican Republic}.
\newblock


\bibitem[Levine et~al\mbox{.}(2020)]%
        {levine2020offline}
\bibfield{author}{\bibinfo{person}{Sergey Levine}, \bibinfo{person}{Aviral
  Kumar}, \bibinfo{person}{George Tucker}, {and} \bibinfo{person}{Justin Fu}.}
  \bibinfo{year}{2020}\natexlab{}.
\newblock \showarticletitle{Offline reinforcement learning: Tutorial, review,
  and perspectives on open problems}.
\newblock \bibinfo{journal}{\emph{arXiv preprint arXiv:2005.01643}}
  (\bibinfo{year}{2020}).
\newblock


\bibitem[Lewis et~al\mbox{.}(2020)]%
        {lewis2020retrieval}
\bibfield{author}{\bibinfo{person}{Patrick Lewis}, \bibinfo{person}{Ethan
  Perez}, \bibinfo{person}{Aleksandra Piktus}, \bibinfo{person}{Fabio Petroni},
  \bibinfo{person}{Vladimir Karpukhin}, \bibinfo{person}{Naman Goyal},
  \bibinfo{person}{Heinrich K{\"u}ttler}, \bibinfo{person}{Mike Lewis},
  \bibinfo{person}{Wen-tau Yih}, \bibinfo{person}{Tim Rockt{\"a}schel},
  {et~al\mbox{.}}} \bibinfo{year}{2020}\natexlab{}.
\newblock \showarticletitle{Retrieval-augmented generation for
  knowledge-intensive nlp tasks}.
\newblock \bibinfo{journal}{\emph{NeurIPS}}  \bibinfo{volume}{33}
  (\bibinfo{year}{2020}), \bibinfo{pages}{9459--9474}.
\newblock


\bibitem[Li et~al\mbox{.}(2018)]%
        {li2018transfer}
\bibfield{author}{\bibinfo{person}{Juncen Li}, \bibinfo{person}{Robin Jia},
  \bibinfo{person}{He He}, {and} \bibinfo{person}{Percy Liang}.}
  \bibinfo{year}{2018}\natexlab{}.
\newblock \showarticletitle{Delete, Retrieve, Generate: A Simple Approach to
  Sentiment and Style Transfer}. In \bibinfo{booktitle}{\emph{North American
  Association for Computational Linguistics (NAACL)}}.
\newblock
\urldef\tempurl%
\url{https://nlp.stanford.edu/pubs/li2018transfer.pdf}
\showURL{%
\tempurl}


\bibitem[Li et~al\mbox{.}(2021)]%
        {li2021focal}
\bibfield{author}{\bibinfo{person}{Lanqing Li}, \bibinfo{person}{Rui Yang},
  {and} \bibinfo{person}{Dijun Luo}.} \bibinfo{year}{2021}\natexlab{}.
\newblock \showarticletitle{{FOCAL}: Efficient Fully-Offline Meta-Reinforcement
  Learning via Distance Metric Learning and Behavior Regularization}. In
  \bibinfo{booktitle}{\emph{ICRL}}.
\newblock
\urldef\tempurl%
\url{https://openreview.net/forum?id=8cpHIfgY4Dj}
\showURL{%
\tempurl}


\bibitem[Lin et~al\mbox{.}(2023)]%
        {lin2023transformers}
\bibfield{author}{\bibinfo{person}{Licong Lin}, \bibinfo{person}{Yu Bai}, {and}
  \bibinfo{person}{Song Mei}.} \bibinfo{year}{2023}\natexlab{}.
\newblock \showarticletitle{Transformers as decision makers: Provable
  in-context reinforcement learning via supervised pretraining}.
\newblock \bibinfo{journal}{\emph{arXiv preprint arXiv:2310.08566}}
  (\bibinfo{year}{2023}).
\newblock


\bibitem[Mitchell et~al\mbox{.}(2021)]%
        {mitchell2021offline}
\bibfield{author}{\bibinfo{person}{Eric Mitchell}, \bibinfo{person}{Rafael
  Rafailov}, \bibinfo{person}{Xue~Bin Peng}, \bibinfo{person}{Sergey Levine},
  {and} \bibinfo{person}{Chelsea Finn}.} \bibinfo{year}{2021}\natexlab{}.
\newblock \showarticletitle{Offline meta-reinforcement learning with advantage
  weighting}. In \bibinfo{booktitle}{\emph{ICML}}. PMLR,
  \bibinfo{pages}{7780--7791}.
\newblock


\bibitem[Munkhdalai and Yu(2017)]%
        {munkhdalai2017meta}
\bibfield{author}{\bibinfo{person}{Tsendsuren Munkhdalai} {and}
  \bibinfo{person}{Hong Yu}.} \bibinfo{year}{2017}\natexlab{}.
\newblock \showarticletitle{Meta networks}. In
  \bibinfo{booktitle}{\emph{ICML}}. PMLR, \bibinfo{pages}{2554--2563}.
\newblock


\bibitem[Nachum et~al\mbox{.}(2018)]%
        {nachum2018data}
\bibfield{author}{\bibinfo{person}{Ofir Nachum},
  \bibinfo{person}{Shixiang~Shane Gu}, \bibinfo{person}{Honglak Lee}, {and}
  \bibinfo{person}{Sergey Levine}.} \bibinfo{year}{2018}\natexlab{}.
\newblock \showarticletitle{Data-efficient hierarchical reinforcement
  learning}.
\newblock \bibinfo{journal}{\emph{NeurIPS}}  \bibinfo{volume}{31}
  (\bibinfo{year}{2018}).
\newblock


\bibitem[Ni et~al\mbox{.}(2023)]%
        {ni2023metadiffuser}
\bibfield{author}{\bibinfo{person}{Fei Ni}, \bibinfo{person}{Jianye Hao},
  \bibinfo{person}{Yao Mu}, \bibinfo{person}{Yifu Yuan}, \bibinfo{person}{Yan
  Zheng}, \bibinfo{person}{Bin Wang}, {and} \bibinfo{person}{Zhixuan Liang}.}
  \bibinfo{year}{2023}\natexlab{}.
\newblock \showarticletitle{Metadiffuser: Diffusion model as conditional
  planner for offline meta-rl}. In \bibinfo{booktitle}{\emph{International
  Conference on Machine Learning}}. PMLR, \bibinfo{pages}{26087--26105}.
\newblock


\bibitem[Paster et~al\mbox{.}(2022)]%
        {paster2022you}
\bibfield{author}{\bibinfo{person}{Keiran Paster}, \bibinfo{person}{Sheila
  Mcilraith}, {and} \bibinfo{person}{Jimmy Ba}.}
  \bibinfo{year}{2022}\natexlab{}.
\newblock \showarticletitle{You can’t count on luck: Why decision
  transformers and rvs fail in stochastic environments}.
\newblock \bibinfo{journal}{\emph{NeurIPS}}  \bibinfo{volume}{35}
  (\bibinfo{year}{2022}), \bibinfo{pages}{38966--38979}.
\newblock


\bibitem[Peng et~al\mbox{.}(2019)]%
        {peng2019text}
\bibfield{author}{\bibinfo{person}{Hao Peng}, \bibinfo{person}{Ankur~P Parikh},
  \bibinfo{person}{Manaal Faruqui}, \bibinfo{person}{Bhuwan Dhingra}, {and}
  \bibinfo{person}{Dipanjan Das}.} \bibinfo{year}{2019}\natexlab{}.
\newblock \showarticletitle{Text generation with exemplar-based adaptive
  decoding}.
\newblock \bibinfo{journal}{\emph{arXiv preprint arXiv:1904.04428}}
  (\bibinfo{year}{2019}).
\newblock


\bibitem[Prudencio et~al\mbox{.}(2023)]%
        {prudencio2023survey}
\bibfield{author}{\bibinfo{person}{Rafael~Figueiredo Prudencio},
  \bibinfo{person}{Marcos~ROA Maximo}, {and} \bibinfo{person}{Esther~Luna
  Colombini}.} \bibinfo{year}{2023}\natexlab{}.
\newblock \showarticletitle{A survey on offline reinforcement learning:
  Taxonomy, review, and open problems}.
\newblock \bibinfo{journal}{\emph{IEEE Transactions on Neural Networks and
  Learning Systems}} (\bibinfo{year}{2023}).
\newblock


\bibitem[Radford et~al\mbox{.}(2019)]%
        {radford2019language}
\bibfield{author}{\bibinfo{person}{Alec Radford}, \bibinfo{person}{Jeffrey Wu},
  \bibinfo{person}{Rewon Child}, \bibinfo{person}{David Luan},
  \bibinfo{person}{Dario Amodei}, \bibinfo{person}{Ilya Sutskever},
  {et~al\mbox{.}}} \bibinfo{year}{2019}\natexlab{}.
\newblock \showarticletitle{Language models are unsupervised multitask
  learners}.
\newblock \bibinfo{journal}{\emph{OpenAI blog}} \bibinfo{volume}{1},
  \bibinfo{number}{8} (\bibinfo{year}{2019}), \bibinfo{pages}{9}.
\newblock


\bibitem[Reed et~al\mbox{.}(2022)]%
        {reed2022generalist}
\bibfield{author}{\bibinfo{person}{Scott Reed}, \bibinfo{person}{Konrad Zolna},
  \bibinfo{person}{Emilio Parisotto}, \bibinfo{person}{Sergio~Gomez
  Colmenarejo}, \bibinfo{person}{Alexander Novikov}, \bibinfo{person}{Gabriel
  Barth-Maron}, \bibinfo{person}{Mai Gimenez}, \bibinfo{person}{Yury Sulsky},
  \bibinfo{person}{Jackie Kay}, \bibinfo{person}{Jost~Tobias Springenberg},
  {et~al\mbox{.}}} \bibinfo{year}{2022}\natexlab{}.
\newblock \showarticletitle{A generalist agent}.
\newblock \bibinfo{journal}{\emph{arXiv preprint arXiv:2205.06175}}
  (\bibinfo{year}{2022}).
\newblock


\bibitem[Rothfuss et~al\mbox{.}(2018)]%
        {rothfuss2018promp}
\bibfield{author}{\bibinfo{person}{Jonas Rothfuss}, \bibinfo{person}{Dennis
  Lee}, \bibinfo{person}{Ignasi Clavera}, \bibinfo{person}{Tamim Asfour}, {and}
  \bibinfo{person}{Pieter Abbeel}.} \bibinfo{year}{2018}\natexlab{}.
\newblock \showarticletitle{Promp: Proximal meta-policy search}.
\newblock \bibinfo{journal}{\emph{arXiv preprint arXiv:1810.06784}}
  (\bibinfo{year}{2018}).
\newblock


\bibitem[Schweighofer et~al\mbox{.}(2021)]%
        {schweighofer2021understanding}
\bibfield{author}{\bibinfo{person}{Kajetan Schweighofer},
  \bibinfo{person}{Markus Hofmarcher}, \bibinfo{person}{Marius-Constantin
  Dinu}, \bibinfo{person}{Philipp Renz}, \bibinfo{person}{Angela
  Bitto-Nemling}, \bibinfo{person}{Vihang~Prakash Patil}, {and}
  \bibinfo{person}{Sepp Hochreiter}.} \bibinfo{year}{2021}\natexlab{}.
\newblock \showarticletitle{Understanding the effects of dataset
  characteristics on offline reinforcement learning}. In
  \bibinfo{booktitle}{\emph{Deep RL Workshop NeurIPS 2021}}.
\newblock


\bibitem[Shen et~al\mbox{.}(2024)]%
        {shen2024hugginggpt}
\bibfield{author}{\bibinfo{person}{Yongliang Shen}, \bibinfo{person}{Kaitao
  Song}, \bibinfo{person}{Xu Tan}, \bibinfo{person}{Dongsheng Li},
  \bibinfo{person}{Weiming Lu}, {and} \bibinfo{person}{Yueting Zhuang}.}
  \bibinfo{year}{2024}\natexlab{}.
\newblock \showarticletitle{Hugginggpt: Solving ai tasks with chatgpt and its
  friends in hugging face}.
\newblock \bibinfo{journal}{\emph{Advances in Neural Information Processing
  Systems}}  \bibinfo{volume}{36} (\bibinfo{year}{2024}).
\newblock


\bibitem[Sodhani et~al\mbox{.}(2021)]%
        {sodhani2021multi}
\bibfield{author}{\bibinfo{person}{Shagun Sodhani}, \bibinfo{person}{Amy
  Zhang}, {and} \bibinfo{person}{Joelle Pineau}.}
  \bibinfo{year}{2021}\natexlab{}.
\newblock \showarticletitle{Multi-task reinforcement learning with
  context-based representations}. In \bibinfo{booktitle}{\emph{ICML}}. PMLR,
  \bibinfo{pages}{9767--9779}.
\newblock


\bibitem[Sutton and Barto(2018)]%
        {sutton2018reinforcement}
\bibfield{author}{\bibinfo{person}{Richard~S Sutton} {and}
  \bibinfo{person}{Andrew~G Barto}.} \bibinfo{year}{2018}\natexlab{}.
\newblock \bibinfo{booktitle}{\emph{Reinforcement learning: An introduction}}.
\newblock \bibinfo{publisher}{MIT press}.
\newblock


\bibitem[Tang et~al\mbox{.}(2023)]%
        {tang2023zeroth}
\bibfield{author}{\bibinfo{person}{Zhiwei Tang}, \bibinfo{person}{Dmitry
  Rybin}, {and} \bibinfo{person}{Tsung-Hui Chang}.}
  \bibinfo{year}{2023}\natexlab{}.
\newblock \showarticletitle{Zeroth-Order Optimization Meets Human Feedback:
  Provable Learning via Ranking Oracles}.
\newblock \bibinfo{journal}{\emph{arXiv preprint arXiv:2303.03751}}
  (\bibinfo{year}{2023}).
\newblock


\bibitem[Todorov et~al\mbox{.}(2012)]%
        {todorov2012mujoco}
\bibfield{author}{\bibinfo{person}{Emanuel Todorov}, \bibinfo{person}{Tom
  Erez}, {and} \bibinfo{person}{Yuval Tassa}.} \bibinfo{year}{2012}\natexlab{}.
\newblock \showarticletitle{MuJoCo: A physics engine for model-based control}.
  In \bibinfo{booktitle}{\emph{2012 IEEE/RSJ IROS}}. IEEE,
  \bibinfo{pages}{5026--5033}.
\newblock
\urldef\tempurl%
\url{https://doi.org/10.1109/IROS.2012.6386109}
\showDOI{\tempurl}


\bibitem[Vezhnevets et~al\mbox{.}(2017)]%
        {vezhnevets2017feudal}
\bibfield{author}{\bibinfo{person}{Alexander~Sasha Vezhnevets},
  \bibinfo{person}{Simon Osindero}, \bibinfo{person}{Tom Schaul},
  \bibinfo{person}{Nicolas Heess}, \bibinfo{person}{Max Jaderberg},
  \bibinfo{person}{David Silver}, {and} \bibinfo{person}{Koray Kavukcuoglu}.}
  \bibinfo{year}{2017}\natexlab{}.
\newblock \showarticletitle{Feudal networks for hierarchical reinforcement
  learning}. In \bibinfo{booktitle}{\emph{ICML}}. PMLR,
  \bibinfo{pages}{3540--3549}.
\newblock


\bibitem[Wang et~al\mbox{.}(2022b)]%
        {wang2022bootstrapped}
\bibfield{author}{\bibinfo{person}{Kerong Wang}, \bibinfo{person}{Hanye Zhao},
  \bibinfo{person}{Xufang Luo}, \bibinfo{person}{Kan Ren},
  \bibinfo{person}{Weinan Zhang}, {and} \bibinfo{person}{Dongsheng Li}.}
  \bibinfo{year}{2022}\natexlab{b}.
\newblock \showarticletitle{Bootstrapped transformer for offline reinforcement
  learning}.
\newblock \bibinfo{journal}{\emph{NeurIPS}}  \bibinfo{volume}{35}
  (\bibinfo{year}{2022}), \bibinfo{pages}{34748--34761}.
\newblock


\bibitem[Wang et~al\mbox{.}(2023)]%
        {wang2023a}
\bibfield{author}{\bibinfo{person}{Yiqi Wang}, \bibinfo{person}{Mengdi Xu},
  \bibinfo{person}{Laixi Shi}, {and} \bibinfo{person}{Yuejie Chi}.}
  \bibinfo{year}{2023}\natexlab{}.
\newblock \showarticletitle{A Trajectory is Worth Three Sentences: Multimodal
  Transformer for Offline Reinforcement Learning}. In
  \bibinfo{booktitle}{\emph{UAI}}.
\newblock
\urldef\tempurl%
\url{https://openreview.net/forum?id=yE1_GpmDOPL}
\showURL{%
\tempurl}


\bibitem[Wang et~al\mbox{.}(2022a)]%
        {wang2022stmaml}
\bibfield{author}{\bibinfo{person}{Zhe Wang}, \bibinfo{person}{Jake Grigsby},
  \bibinfo{person}{Arshdeep Sekhon}, {and} \bibinfo{person}{Yanjun Qi}.}
  \bibinfo{year}{2022}\natexlab{a}.
\newblock \showarticletitle{{ST}-{MAML}: A Stochastic-Task based Method for
  Task-Heterogeneous Meta-Learning}. In \bibinfo{booktitle}{\emph{UAI}}.
\newblock
\urldef\tempurl%
\url{https://openreview.net/forum?id=rrlMyPUs9gc}
\showURL{%
\tempurl}


\bibitem[Xu et~al\mbox{.}(2022)]%
        {xu2022prompting}
\bibfield{author}{\bibinfo{person}{Mengdi Xu}, \bibinfo{person}{Yikang Shen},
  \bibinfo{person}{Shun Zhang}, \bibinfo{person}{Yuchen Lu},
  \bibinfo{person}{Ding Zhao}, \bibinfo{person}{Joshua Tenenbaum}, {and}
  \bibinfo{person}{Chuang Gan}.} \bibinfo{year}{2022}\natexlab{}.
\newblock \showarticletitle{Prompting Decision Transformer for Few-Shot Policy
  Generalization}. In \bibinfo{booktitle}{\emph{ICML}}. PMLR,
  \bibinfo{pages}{24631--24645}.
\newblock


\bibitem[Yao et~al\mbox{.}(2022)]%
        {yao2022react}
\bibfield{author}{\bibinfo{person}{Shunyu Yao}, \bibinfo{person}{Jeffrey Zhao},
  \bibinfo{person}{Dian Yu}, \bibinfo{person}{Nan Du}, \bibinfo{person}{Izhak
  Shafran}, \bibinfo{person}{Karthik Narasimhan}, {and} \bibinfo{person}{Yuan
  Cao}.} \bibinfo{year}{2022}\natexlab{}.
\newblock \showarticletitle{React: Synergizing reasoning and acting in language
  models}.
\newblock \bibinfo{journal}{\emph{arXiv preprint arXiv:2210.03629}}
  (\bibinfo{year}{2022}).
\newblock


\bibitem[Yu et~al\mbox{.}(2020a)]%
        {yu2020gradient}
\bibfield{author}{\bibinfo{person}{Tianhe Yu}, \bibinfo{person}{Saurabh Kumar},
  \bibinfo{person}{Abhishek Gupta}, \bibinfo{person}{Sergey Levine},
  \bibinfo{person}{Karol Hausman}, {and} \bibinfo{person}{Chelsea Finn}.}
  \bibinfo{year}{2020}\natexlab{a}.
\newblock \showarticletitle{Gradient surgery for multi-task learning}.
\newblock \bibinfo{journal}{\emph{NeurIPS}}  \bibinfo{volume}{33}
  (\bibinfo{year}{2020}), \bibinfo{pages}{5824--5836}.
\newblock


\bibitem[Yu et~al\mbox{.}(2020b)]%
        {yu2020meta}
\bibfield{author}{\bibinfo{person}{Tianhe Yu}, \bibinfo{person}{Deirdre
  Quillen}, \bibinfo{person}{Zhanpeng He}, \bibinfo{person}{Ryan Julian},
  \bibinfo{person}{Karol Hausman}, \bibinfo{person}{Chelsea Finn}, {and}
  \bibinfo{person}{Sergey Levine}.} \bibinfo{year}{2020}\natexlab{b}.
\newblock \showarticletitle{Meta-world: A benchmark and evaluation for
  multi-task and meta reinforcement learning}. In
  \bibinfo{booktitle}{\emph{Conference on robot learning}}. PMLR,
  \bibinfo{pages}{1094--1100}.
\newblock


\bibitem[Zaheer et~al\mbox{.}(2017)]%
        {zaheer2017deep}
\bibfield{author}{\bibinfo{person}{Manzil Zaheer}, \bibinfo{person}{Satwik
  Kottur}, \bibinfo{person}{Siamak Ravanbakhsh}, \bibinfo{person}{Barnabas
  Poczos}, \bibinfo{person}{Russ~R Salakhutdinov}, {and}
  \bibinfo{person}{Alexander~J Smola}.} \bibinfo{year}{2017}\natexlab{}.
\newblock \showarticletitle{Deep sets}.
\newblock \bibinfo{journal}{\emph{NeurIPS}}  \bibinfo{volume}{30}
  (\bibinfo{year}{2017}).
\newblock


\bibitem[Zhang et~al\mbox{.}(2020)]%
        {zhang2020learning}
\bibfield{author}{\bibinfo{person}{Amy Zhang}, \bibinfo{person}{Shagun
  Sodhani}, \bibinfo{person}{Khimya Khetarpal}, {and} \bibinfo{person}{Joelle
  Pineau}.} \bibinfo{year}{2020}\natexlab{}.
\newblock \showarticletitle{Learning robust state abstractions for
  hidden-parameter block MDPs}.
\newblock \bibinfo{journal}{\emph{arXiv preprint arXiv:2007.07206}}
  (\bibinfo{year}{2020}).
\newblock


\bibitem[Zhou et~al\mbox{.}(2022)]%
        {zhou2022unsupervised}
\bibfield{author}{\bibinfo{person}{Mingyang Zhou}, \bibinfo{person}{Licheng
  Yu}, \bibinfo{person}{Amanpreet Singh}, \bibinfo{person}{Mengjiao Wang},
  \bibinfo{person}{Zhou Yu}, {and} \bibinfo{person}{Ning Zhang}.}
  \bibinfo{year}{2022}\natexlab{}.
\newblock \showarticletitle{Unsupervised vision-and-language pre-training via
  retrieval-based multi-granular alignment}. In
  \bibinfo{booktitle}{\emph{CVPR}}. \bibinfo{pages}{16485--16494}.
\newblock


\bibitem[Zhu et~al\mbox{.}(2019)]%
        {zhu2019episodic}
\bibfield{author}{\bibinfo{person}{Guangxiang Zhu}, \bibinfo{person}{Zichuan
  Lin}, \bibinfo{person}{Guangwen Yang}, {and} \bibinfo{person}{Chongjie
  Zhang}.} \bibinfo{year}{2019}\natexlab{}.
\newblock \showarticletitle{Episodic reinforcement learning with associative
  memory}. In \bibinfo{booktitle}{\emph{ICLR}}.
\newblock


\end{thebibliography}

\appendix

\end{document}